\documentclass{article} %
\usepackage{style/iclr2023_conference,times}

\usepackage{graphicx}

\usepackage{amsmath,amsfonts,bm}

\def\eqref#1{equation~\ref{#1}}
\def\Eqref#1{Equation~\ref{#1}}

\def\1{\bm{1}}

\DeclareMathAlphabet{\mathsfit}{\encodingdefault}{\sfdefault}{m}{sl}
\SetMathAlphabet{\mathsfit}{bold}{\encodingdefault}{\sfdefault}{bx}{n}

\usepackage{booktabs}
\usepackage{placeins}
\usepackage[pagebackref=true]{hyperref} 
\renewcommand*\backref[1]{\ifx#1\relax \else (Cited on page #1) \fi}
\usepackage[normalem]{ulem}
\usepackage{url}
\usepackage{tcolorbox}
\usepackage{subcaption}
\renewcommand{\headrulewidth}{0pt}
\usepackage{adjustbox}

\usepackage{enumitem}
\usepackage{cleveref}
\usepackage{xcolor}
\hypersetup{
    colorlinks,
    linkcolor={red!50!black},
    citecolor={blue!50!black},
    urlcolor={blue!80!black}
}

\usepackage{listings}
\lstdefinestyle{mystyle}{
    language=Python,
    basicstyle=\ttfamily\footnotesize,
    commentstyle=\color{olive},
    keywordstyle=\color{blue},
    numberstyle=\tiny\color{gray},
    stringstyle=\color{purple},
    breakatwhitespace=false,         
    breaklines=true,                 
    captionpos=b,                    
    keepspaces=true,                 
    numbers=left,                    
    numbersep=5pt,                  
    showspaces=false,
    showstringspaces=false,
    showtabs=false,                  
    tabsize=2
}
\definecolor{magenta}{RGB}{255,0,255}
\definecolor{maroon}{RGB}{128,0,0}
\definecolor{gold}{RGB}{255,215,0}
\definecolor{peach}{RGB}{255,229,180}
\definecolor{darkpeach}{RGB}{255,180,120}
\definecolor{cerulean}{RGB}{0,123,167}
\definecolor{salmon}{RGB}{250,128,114}

\title{Weak-to-Strong Generalization: Eliciting Strong Capabilities With Weak Supervision}

\author{
  Collin Burns\thanks{Primary authors. This was a joint project of the Superalignment Generalization team. Correspondence to \texttt{generalization@openai.com}.
  Code is available at \href{https://github.com/openai/weak-to-strong}{\url{github.com/openai/weak-to-strong}}.
  } 
  \And 
  Pavel Izmailov$^*$
  \And 
  Jan Hendrik Kirchner$^*$  
  \And
  Bowen Baker$^*$
  \And 
  Leo Gao$^*$
  \AND 
  Leopold Aschenbrenner$^*$
  \And
  Yining Chen$^*$
  \And
  Adrien Ecoffet$^*$
  \And
  Manas Joglekar$^*$
  \AND
  Jan Leike
  \And
  Ilya Sutskever
  \And
  Jeff Wu$^*$
  \AND
  \normalfont{OpenAI}
}

\iclrfinalcopy %
\begin{document}

\maketitle

\begin{abstract}

Widely used alignment techniques, such as reinforcement learning from human feedback (RLHF), rely on the ability of humans to supervise model behavior---for example, to evaluate whether a model faithfully followed instructions or generated safe outputs.
However, future superhuman models will behave in complex ways too difficult for humans to reliably evaluate; humans will only be able to \textit{weakly supervise} superhuman models.
We study an analogy to this problem: can weak model supervision elicit the full capabilities of a much stronger model?
We test this using a range of pretrained language models in the GPT-4 family on natural language processing (NLP), chess, and reward modeling tasks. 
We find that when we naively finetune strong pretrained models on labels generated by a weak model, they consistently perform better than their weak supervisors, a phenomenon we call \textit{weak-to-strong generalization}.
However, we are still far from recovering the full capabilities of strong models with naive finetuning alone, suggesting that techniques like RLHF may scale poorly to superhuman models without further work. 
We find that simple methods can often significantly improve weak-to-strong generalization: for example, when finetuning GPT-4 with a GPT-2-level supervisor and an auxiliary confidence loss, we can recover close to GPT-3.5-level performance on NLP tasks.
Our results suggest that it is feasible to make empirical progress today on a fundamental challenge of aligning superhuman models.

\end{abstract}

\section{Introduction}

We mainly steer or \textit{align} today's models with reinforcement learning from human feedback~(RLHF): we reinforce behaviors that human evaluators rate highly and penalize behaviors that evaluators rate poorly~\citep{christiano2017deep,stiennon2020learning,ouyang2022training,glaese2022improving,bai2022training}.
This procedure is very effective when human evaluators can tell if model behavior is good or bad and is a core part of training modern language model assistants such as ChatGPT. 

However, superhuman models will be capable of complex and creative behaviors that humans cannot fully understand. For example, if a superhuman assistant model generates a million lines of extremely complicated code, humans will not be able to provide reliable supervision for key alignment-relevant tasks, including: whether the code follows the user's intentions, whether the assistant model answers questions about the code honestly, whether the code is safe or dangerous to execute, and so on.
As a result, if we finetune a superhuman model with human supervision on a reward modeling (RM) or safety classification task, it is unclear how that model will generalize to complicated behaviors that humans could not reliably supervise themselves.

This leads to a fundamental technical challenge of aligning superhuman models (superalignment): how can weak supervisors control models much smarter than them?
Despite the importance of this problem, it is difficult to empirically study today.
Most prior work on alignment has either confronted this core challenge head-on---but been restricted to primarily theoretical frameworks and toy problems~\citep{irving2018ai,christiano2018supervising,leike2018scalable,demski2019embedded,hubinger2019risks}, or empirically studied humans supervising today's models---without addressing the core challenges that may arise with superhuman models~\citep{christiano2017deep, wu2021recursively,ouyang2022training,bowman2022measuring,saunders2022self}. 
In contrast, we would ideally like to have a setup that captures core challenges of aligning future superhuman models while \textit{also} being able to make iterative empirical progress today. 

We propose a simple setup for studying the problem of humans supervising superhuman models by considering an analogy: can we use \textit{weak models} to supervise \textit{strong models}? 
We can empirically test this by finetuning large (strong) pretrained models on labels generated by small (weak) models and observing how they generalize. 
Just like the problem of humans supervising superhuman models, our setup is an instance of what we call the \textit{weak-to-strong learning} problem.

Why should weak-to-strong learning be possible?
On the one hand, the strong model could simply learn to imitate the weak supervisor, including its errors, since that is what we would naively train it to do. 
On the other hand, strong pretrained models should already have good representations of the alignment-relevant tasks we care about. 
For example, if a model can generate complicated code, then it should intuitively also know whether that code faithfully adheres to the user's instructions.
As a result, for the purposes of alignment we do not need the weak supervisor to teach the strong model new capabilities; instead, we simply need the weak supervisor to elicit what the strong model \textit{already knows}. 
This gives us hope that the strong model can generalize beyond the weak supervision, solving even hard problems for which the weak supervisor can only give incomplete or flawed training labels. We call this phenomenon \textit{weak-to-strong generalization}.

\begin{figure}[t]
\centering
\includegraphics[width=1.0 \textwidth]{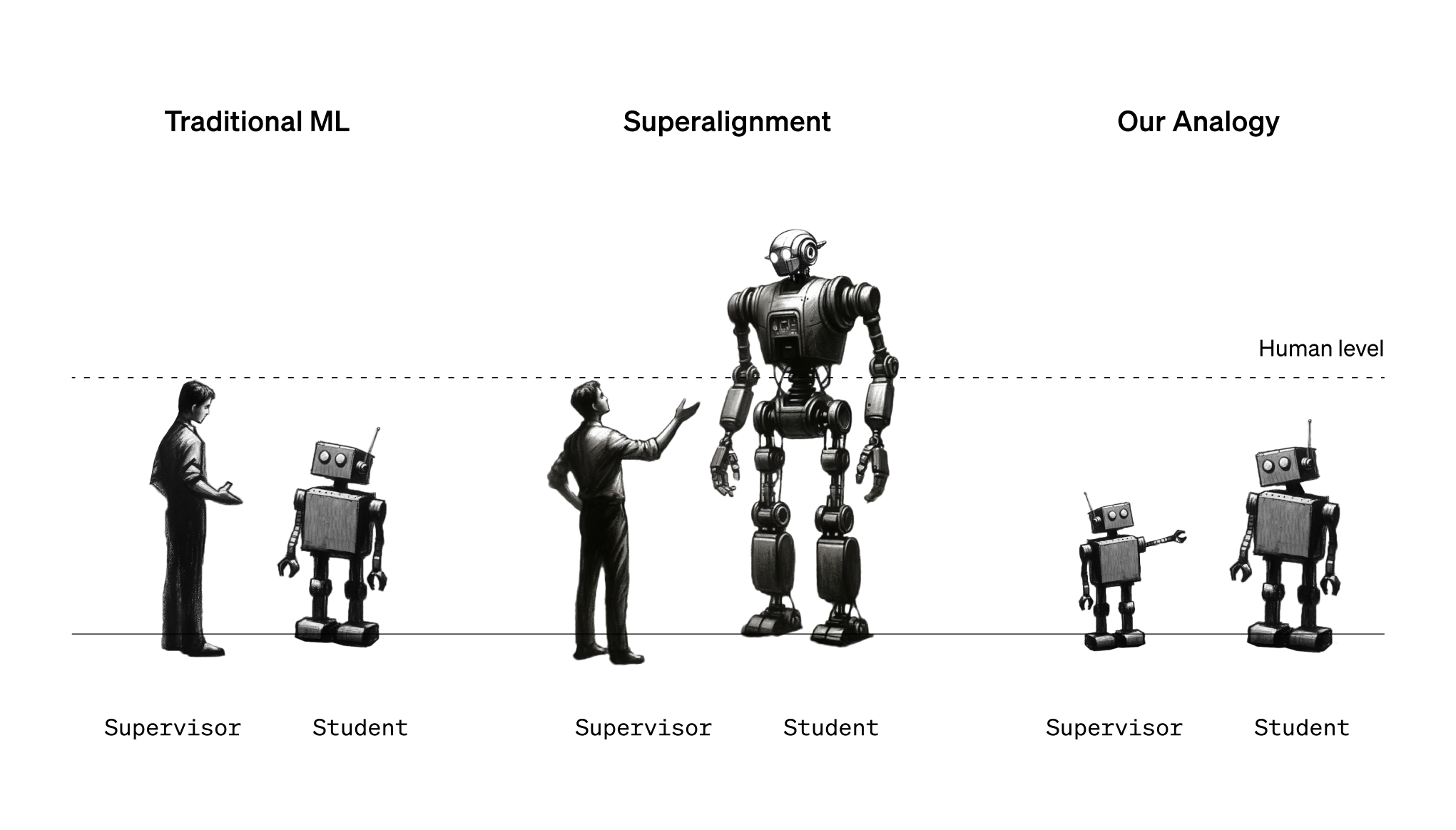}
\caption{\textbf{An illustration of our methodology.} Traditional ML focuses on the setting where humans supervise models that are weaker than humans. For the ultimate superalignment problem, humans will have to supervise models much smarter than them. We study an analogous problem today: using weak models to supervise strong models.}
\label{fig:methodology}
\vspace{-10pt}
\end{figure}

We study our weak-to-strong learning setup (\Cref{sec:methodology}) by finetuning base (i.e.\ pretrained-only) language models from the GPT-4 family~\citep{openai2023gpt},\footnote{These models share the same general architecture and pretraining dataset as GPT-4. However, this model series does not include the models known as GPT-2, GPT-3, and GPT-3.5.} spanning 7 orders of magnitude (OOMs) of pretraining compute, across three settings: a large set of popular natural language processing (NLP) benchmarks, chess puzzles, and our internal ChatGPT reward modeling dataset.
Our main findings include:

\begin{figure}
    \centering
    \includegraphics[width=\linewidth]{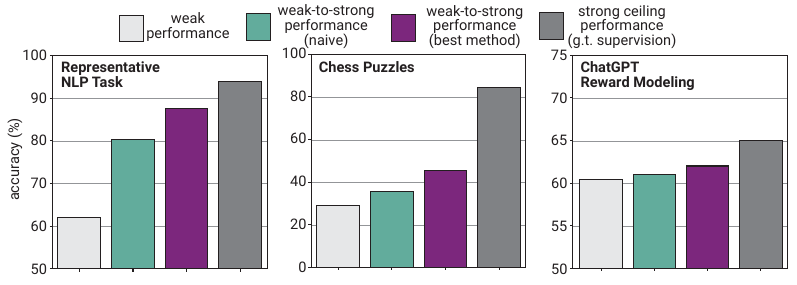}
    \caption{\textbf{Strong models trained with weak supervision generalize beyond their supervisor, and improving weak-to-strong generalization is tractable.}
    We show test accuracy on a representative NLP task (left), chess puzzles (middle) and the ChatGPT reward modeling task (right). We show the weak supervisor trained on ground truth labels (light grey) and the strong student trained with weak supervision naively (green), with the best method in each setting (purple), or with ground truth supervision (dark grey). 
    For NLP and chess we supervise GPT-4 using GPT-2-level supervision, while for reward modeling we supervise a 3.5-level model using GPT-2-level supervision.
    The best method is the auxiliary confidence loss for the NLP task (\Cref{sec:auxloss}), bootstrapping for Chess puzzles (\Cref{sec:bootstrapping}), and unsupervised generative finetuning for reward modeling (\Cref{sec:rm-sft}; generative-finetuning is also used for the strong ceiling performance).  
    }
    \label{fig:accuracy-gap}
\end{figure}

\begin{enumerate}

\item \textbf{Strong pretrained models naturally generalize beyond their weak supervisors.} If we naively finetune strong models with labels generated by weak models, they consistently outperform their weak supervisors (\Cref{section:baselines}). For example, on NLP tasks, if we finetune GPT-4 with labels from a GPT-2-level model, we typically recover about half of the performance gap between the two models.

\item \textbf{Naively finetuning on weak supervison is not enough.} Despite positive weak-to-strong generalization, there still remains a substantial gap between strong models finetuned with weak supervision and strong models finetuned with ground truth supervision.
Weak-to-strong generalization is particularly poor for ChatGPT reward modeling. 
Collectively, our results provide empirical evidence that naive RLHF will likely scale poorly to superhuman models without additional work.

\item \textbf{Improving weak-to-strong generalization is tractable.} 
We find that we can improve performance by encouraging strong models to have confident predictions with an auxiliary loss, bootstrapping supervision with intermediate models, and improving model representations with unsupervised finetuning. For example, when supervising GPT-4 with a GPT-2-level model on NLP tasks using the auxiliary confidence loss, we typically recover nearly 80\% of the performance gap between the weak and strong models.

\end{enumerate}

Our work has important limitations. 
None of our methods work consistently in all settings, and especially in the RM setting we are still far from recovering the full performance gap between weak and strong models. Thus our methods serve more as proofs-of-concept that weak-to-strong generalization is tractable, rather than practical solutions we recommend deploying today.
Furthermore, there are still important disanalogies between our empirical setup and aligning superhuman models that we did not address (\Cref{sec:discussion}); continuously refining our basic setup will be important for ensuring that research today continues to make real progress toward aligning the superhuman models we develop in the future.

Despite the limitations of our work, we find our results to be highly encouraging. 
We show that substantial weak-to-strong generalization is not only possible, but actually a widespread phenomenon. 
We also show that with very simple methods, we can drastically improve the ability of weak supervisors to elicit knowledge from strong models.
With much more progress in this direction, we could get to the point where we can use weak supervisors to reliably elicit knowledge from much stronger models, at least for some key tasks that we care about.
This may allow us to develop superhuman reward models or safety classifiers, which we could in turn use to align superhuman models.

Aligning superhuman models is essential for making them safe; there is increasing recognition that failing to align such powerful models has the potential to be catastrophic, making this one of the most important unsolved technical problems in the world~\citep{ai-risk-open-letter}.
We think it is now more tractable than ever to make rapid iterative empirical progress toward solving this problem.

\section{Related Work}

We study how we can leverage the generalization properties of deep neural networks to solve weak-to-strong learning. 
Our problem setting and methods are closely connected to many existing research areas.

\textbf{Weakly-supervised learning.}\quad 
Weak-to-strong learning is a special type of weakly supervised learning---a setting in which models are trained using unreliable labels~\citep{bach2017learning, ratner2017snorkel, Guo_2018_ECCV}.
There is also a rich literature on the related problem of learning from noisy labels~\citep{song2022learning}. 
Common methods include bootstrapping~\citep{reed2014training, han2018co, li2020dividemix}, noise-robust losses~\citep{zhang2018generalized, hendrycks2018using,ma2020normalized}, and noise modeling~\citep{Yi_2019_CVPR}. 
Unlike most work on label noise, the errors in our weak supervision are much harder to address than uniform label noise, instead having ``instance-dependent'' errors ~\citep{frenay2013classification}.
Semi-supervised learning, in which labels are only available for a subset of the data, is also closely related~\citep{kingma2014semi, laine2016temporal, berthelot2019mixmatch}. We could also study our problem in a semi-supervised setting by having an ``easy'' subset of examples that weak supervisors provide reliable labels for and a subset of unlabeled ``hard'' examples that the weak supervisor can't reliably label, a problem which we call ``easy-to-hard generalization'' (see \Cref{sec:app-easy-hard}).

\textbf{Student-teacher training.}\quad 
The framework of first training a teacher and then training a student on teacher's pseudo-labels is widely used in semi-supervised learning~\citep{laine2016temporal, tarvainen2017mean, xie2020self}, domain adaptation~\citep{french2017self,shu2018dirt}, and knowledge distillation~\citep{hinton2015distilling,gou2021knowledge, stanton2021does, beyer2022knowledge}.
In contrast to most prior work, we focus on the setting where the student is much more capable than the teacher.

\citet{furlanello2018born} and \citet{xie2020self} also consider cases where the student is at least as capable as the teacher.  However in their settings the student is randomly initialized and has access to ground truth labels.
Moreover, compared to most past work we are focused on qualitatively \textit{very} weak supervision. 
For example, we are interested in huge leaps in generalization, similar to going from ``3rd grade-level'' supervisors to ``12th grade-level'' student models. 
Despite these differences with past work, we expect many methods from semi-supervised learning and domain adaptation to translate to our setting. For example, we found that a type of confidence auxiliary loss similar to past work~\citep{grandvalet2004semi} improves weak-to-strong generalization in Section \ref{sec:method}.

\textbf{Robustness of pretraining and finetuning.} \quad 
Many papers have shown that pretraining on massive, diverse data leads to more robust representations that generalize better out-of-distribution~\citep{hendrycks2019using,hendrycks2020pretrained,radford2021learning,liu2022empirical}. 
Finetuning typically improves in-distribution generalization, but often performs poorly out-of-distribution, sometimes even degrading performance relative to zero-shot prompting~\citep{kumar2022fine, wortsman2022robust, awadalla2022exploring}.
Recent approaches to mitigating this problem include weight ensembling~\citep{wortsman2022robust, wortsman2022model}, finetuning only a subset of layers~\citep{kirichenko2023last, lee2022surgical}, or mitigating the distortion effects that finetuning has on pretrained features~\citep{kumar2022fine}.
We did not find strong results in preliminary explorations of approaches similar to these (\Cref{sec:app-other-methods}), but we expect that with more thorough explorations one may be able to attain much stronger results with these or other ideas from the robust finetuning literature.

\textbf{Debiasing.}\quad 
In weak-to-strong generalization, the weak labels contain a specific form of bias, which results from the weak models' lack of capability.
There is a substantial literature on learning from biased training data~\citep{bellamy2018ai}. However, most work focuses on \emph{known} biases, for example where we know that the models perform worse on minority groups. For known biases, common methods include Group Distributionally Robust Optimization~\citep{sagawa2019distributionally}, adversarial training~\citep{zhang2018mitigating}, and model editing~\citep{santurkar2021editing, meng2022locating}. In contrast, our setting can be viewed as a particularly difficult debiasing problem where the bias is unknown. Some methods that automatically discover and mitigate biases include clustering~\citep{sohoni2020no}, loss variance reduction~\citep{khani2019maximum}, and auditing and re-training on high-loss group~\citep{kim2019multiaccuracy, liu2021just}.

\textbf{Imitation and preference learning.}\quad 
The goal of alignment is to steer already-capable models to do what we want them to do.
For example, the base GPT-4 model is good at generating text following its pretraining distribution, but does not readily follow instructions.
To align pretrained language models today, we finetune them using imitation learning on human demonstrations~\citep{bain1995framework,atkeson1997robot} or by using methods such as reinforcement learning from human feedback (RLHF)~\citep{christiano2017deep,stiennon2020learning,ouyang2022training,glaese2022improving,bai2022training}. 
Constitutional AI~\citep{bai2022constitutional, lee2023rlaif} leverages AI feedback to align language models, but still uses an initial RLHF phase.
However, both imitation learning and preference learning assume high-quality human supervision, making it unclear if they will work for superhuman models.

\textbf{Scalable oversight.}\quad 
Scalable oversight techniques aim to improve the ability of humans to supervise models.  
For example, humans may ask models to critique the outputs of other models~\citep{irving2018ai,saunders2022self} or use models to help decompose a problem into simpler sub-problems~\citep{leike2018scalable,christiano2018supervising,lightman2023let}.
Scalable oversight methods typically take advantage of special problem structure, like decomposability or the fact that evaluation is easier than generation. 
In contrast to improving human supervision, we focus on generalizing beyond human supervision such that models perform well even in settings we cannot reliably supervise.
That said, our weak-to-strong learning setup can be used to compare scalable oversight methods, generalization-based methods, and more.
Our setup also resembles a proposal for measuring progress on scalable oversight known as ``sandwiching'', which uses weak and strong humans~\citep{cotra2021sandwiching,bowman2022alignment}.

\textbf{Knowledge elicitation and honesty.}\quad
\citet{christiano2022eliciting} introduced a theoretical problem called Eliciting Latent Knowledge (ELK), in which the goal is to elicit latent knowledge from a superhuman machine learning model even under worst case assumptions.  For example, a special case of ELK is honesty~\citep{evans2021truthful}, where the goal is for the models to report their true beliefs\footnote{Like \cite{evans2021truthful}, we define \textit{honesty} to mean a model reporting what it \textit{believes} to be true, in contrast to \text{truthfulness} which asks whether what a model reports \textit{is} true.}.
\citet{wentworth2020alignment} hypothesizes a tendency for neural networks to develop ``natural abstractions'' that are easier to elicit. Recent empirical work on ELK includes a benchmark for measurement tampering~\citep{roger2023measurement}, methods for discovering latent knowledge~\citep{burns2023discovering}, and studies of honesty
\citep{li2023inference, pacchiardi2023catch}.
Our setting can be viewed as a general methodology for empirically studying problems like ELK and honesty across a wide range of tasks.

\section{Methodology}
\label{sec:methodology}

A core challenge of superalignment is that humans will need to supervise models much smarter
than us. This is a special case of what we call the \textit{weak-to-strong learning problem}: how can a weak
supervisor oversee a model much smarter than it? In this paper, we study a simple analogy, in which we replace the weak human supervisor with a weak model supervisor.

For a given task of interest, consisting of a dataset and a performance metric, we:
\begin{enumerate}
    \item \textbf{Create the weak supervisor.} Throughout most of this work, we create weak supervisors by finetuning small pretrained models on ground truth labels.\footnote{In \Cref{sec:app:more-weak-to-strong} and \Cref{sec:app-simulation} we study other synthetic weak supervisors. Future work could test many more sources of weak supervision, such as by having 3rd grader humans provide labels.}
    We call the performance of the weak supervisor the \textit{weak performance}, and we generate \emph{weak labels} by taking the weak model's predictions on a held-out set of examples.
    \item \textbf{Train a strong student model with weak supervision.} We finetune a strong model with the generated weak labels. We call this model the \emph{strong student model} and its resulting performance the \textit{weak-to-strong performance}.
    \item \textbf{Train a strong model with ground truth labels as a ceiling.} Finally, for comparison, we finetune a strong model with ground truth labels.\footnote{For tasks solved by superhuman models that humans cannot evaluate, we will not have access to ground truth labels. However, we allow access to ground truth labels in our experimental setting today for scientific and evaluation purposes. Note that we evaluated weak-to-strong performance against ground truth many times while iterating on methods; however, we held out our largest model (GPT-4) and about half of NLP tasks throughout the project.
}  We call this model's resulting performance the \textit{strong ceiling performance}. Intuitively, this should correspond to ``everything the strong model knows,'' i.e. the strong model applying its full capabilities to the task.
\end{enumerate}

For more details on how we train each model, see \Cref{app:sec:details}.

Typically, weak-to-strong performance will be between weak performance and strong ceiling performance.  We define the \textbf{performance gap recovered (PGR)} as a function of the above three performances (weak, weak-to-strong, and strong ceiling) as shown in the illustration below.

{
\begin{center}
    \includegraphics[width=0.65\linewidth]{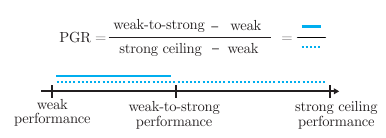}
    \label{fig:AGR-definition}
\end{center}
}

PGR measures the fraction of the performance gap (the difference in performance between the weak and strong ceiling models) that we can recover with weak supervision. If we achieve perfect weak-to-strong generalization, PGR is 1. If the weak-to-strong model does no better than the weak supervisor, then PGR is 0.

\textbf{Advantages.}\quad
Our setup has a number of advantages, including:
\begin{enumerate}
    \item It can be studied with any pair of weak and strong models, making it easy to study scaling laws and not requiring access to expensive state-of-the-art models. Moreover, it does not require working with humans, so feedback loops are fast. 
    \item It can be studied for any task of interest, making it easy to empirically test across a wide range of settings.
    \item Success will be practically useful even before we develop superhuman models: for example, if we find ways to align GPT-4 with only weak human supervision or with only GPT-3-level supervision, that would make it more convenient to align models today.
\end{enumerate}

\textbf{Limitations.}\quad
Our setup still has important disanalogies to the ultimate problem of aligning superhuman models. 
We view our setup as removing one of the main disanalogies in prior work, not as providing a final, perfectly analogous setup. Two remaining disanalogies include:
\begin{enumerate}
    \item \textbf{Imitation saliency.} Future superhuman models will likely have salient representations of human behaviors, but our strong models may not have learned features relevant for imitating weak model predictions; simply imitating the weak supervisor may thus be an easier failure mode to avoid in our setting than it will be in the future. More generally, the types of errors weak models make today may be different from the types of errors humans will make when attempting to supervise superhuman models.%
    \item \textbf{Pretraining leakage.} Our pretraining data implicitly contains supervision from humans. It may thus be artificially easy to elicit strong models' capabilities in our setting, since they were directly pretrained to observe strong (human-level) performance. Superhuman-level performance may not be directly observed in the same way---superhuman knowledge might be more latent, e.g.~because it was learned from self-supervised learning---and thus might be harder to elicit from superhuman models in the future. 
\end{enumerate}

More generally, we do not yet know how superhuman models will be built, but they could develop new inductive biases that are qualitatively different from today's models.
We view iterating on our methodology to produce even more analogous setups as a key priority for future work, as we discuss in more detail in \Cref{sec:discussion}.

\section{Main Results}
\label{sec:results}

In this section, we report our main empirical results, including baselines and promising methods.

\subsection{Tasks}
\label{sec:tasks}

\textbf{Popular natural language processing benchmarks.}\quad
We consider 22 popular NLP classification datasets covering ethics, commonsense reasoning, natural language inference, sentiment analysis, and other domains. 
We convert all datasets to binary classification tasks and approximately balance the classes. 
We produce soft labels from the weak model.  See a full list of the datasets and their sources in \Cref{table:datasets}.

\textbf{Chess puzzles.}\quad
We use the dataset originally introduced in \citet{schwarzschild2021can}, which contains chess puzzles from the \url{lichess.org} website~\citep{lichess}.
Each puzzle consists of a chess position, and a sequence of optimal moves to play to solve the puzzle.
For our evaluation, we predict the first move played, which is the best move in the given chess position.
We illustrate the data format in Appendix Figure \ref{fig:chess-data}.  For weak labels, we sample from the weak model with temperature 0. Note that unlike the other binary classification tasks we study in this paper, this is a generative task.

\textbf{ChatGPT reward modeling.}\quad
The standard approach to aligning models today is reinforcement learning from human feedback (RLHF).
A critical step of RLHF is to train a reward model (RM) to predict human preferences between model responses. 
Specifically, a reward model is trained on a dataset consisting of dialogs between a human and an assistant model. 
For each query, the humans compare multiple possible responses (completions) from the assistant, providing human preference data.
Then, a reward model is trained to predict the results of pairwise comparisons between completions.
Finally, the assistant model is trained by optimizing against the reward model with reinforcement learning (RL).  In our work, we do not study the RL step, and instead assume the goal is to maximize reward model accuracy.
For more details on reward models, see e.g.~\citet{ouyang2022training}.  We use a proprietary dataset used to train ChatGPT reward models.

For more details about our tasks and setup, see \Cref{app:sec:details}.

\subsection{Naively finetuning on weak labels}
\label{section:baselines}

\begin{figure}
    \vspace{-0.85cm}
    \centering
    \includegraphics[width=0.9\linewidth]{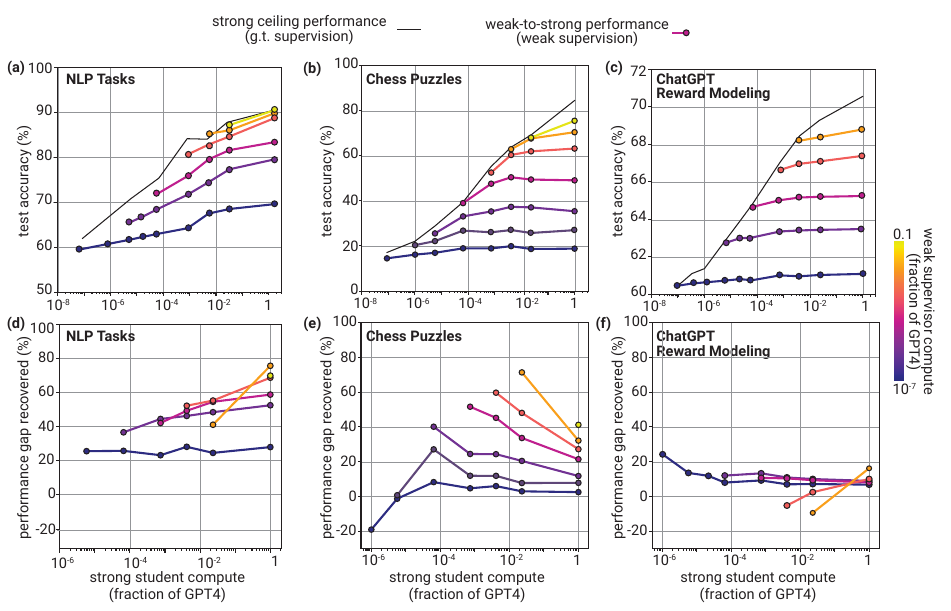}
    \caption{\textbf{Promising weak-to-strong generalization with naive finetuning on NLP tasks and chess, but poor generalization on the ChatGPT reward modeling task.}
    \textbf{(a,b,c)} Test accuracy as a function of strong student size on \textbf{(a)} NLP tasks, \textbf{(b)} chess puzzles, and \textbf{(c)} the ChatGPT reward modeling task. 
 Accuracy of strong students trained with ground truth in black, accuracy of strong students trained with weak supervision shown with colored lines (hue indicates size of weak supervisor). \textbf{(d,e,f)} Same as panels \textbf{a,b,c} but for performance gap recovered (see \Cref{sec:methodology} for details).  For NLP settings, we compute the median across tasks (see \Cref{fig:all_grid_nlp} for full details). 
 We find decent weak-to-strong generalization and even positive PGR scaling on NLP tasks, decent generalization for small supervisor-student gaps but negative PGR scaling on chess puzzles, and both poor generalization and scaling for ChatGPT reward modeling.
    }
    \label{fig:rm-aggreagate}
    \vspace{-0.15cm}
\end{figure}

In each of these 3 settings (NLP tasks, chess puzzles, and reward modeling) we evaluate how well strong students generalize when naively finetuned on labels generated by weak supervisors.
We study pretrained language models from the GPT-4 family~\citep{openai2023gpt}, which allow us to study student-supervisor compute disparities of many orders of magnitude.
We find that PGRs are almost universally positive---in virtually all settings that we studied, and across almost all student and supervisor sizes, students outperform their supervisors (\Cref{fig:rm-aggreagate}).

On the popular NLP benchmarks, we find especially promising weak-to-strong generalization: strong models trained with weak supervision can often generalize to a substantially higher performance than the weak model itself. Even with very weak supervisors and strong models with many orders of magnitude more compute, we recover more than 20\% of the performance gap. 
The PGR increases both with weak supervisor size and with strong student size; 
for the largest students, the PGR is often above 50\%.

We see more mixed results in the chess puzzle setting. 
In particular, when using the smallest weak models, the PGR is close to zero and the test accuracy curves appear flat.
However, as the size of the weak supervisor increases, the PGR increases substantially;  for small supervisor-student gaps, PGR can be above 40\%.
Unlike in the NLP setting, where PGR improves with the strong student size, PGR \textit{decreases} with the strong student size for a given weak supervisor on chess puzzles. The corresponding test accuracy curves appear concave, potentially exhibiting inverse scaling~\citep{mckenzie2023inverse} in strong student size.

Finally, we find that weak-to-strong generalization is poor by default in the ChatGPT reward model setting. We are usually only able to recover roughly 10\% of the performance gap between the weak supervisor and the strong student. Even for relatively small gaps in compute between the weak and strong models, PGR almost never exceeds 20\%.

In general, across all our settings, we observe weak-to-strong generalization: strong students consistently outperform their weak supervisors.
It is not obvious why this should happen at all---especially from naive finetuning alone---and it gives us hope that weak-to-strong learning is a tractable problem.
At the same time, our results suggest that naively using weak, human-level supervision will be insufficient to align strong, superhuman models; we will need qualitatively new techniques to solve superalignment.

\subsection{Improving Weak-to-Strong Generalization is Tractable}
\label{sec:method}

We now show that we can use simple methods to substantially improve weak-to-strong generalization. 
While none of the methods we test works universally, these methods are proofs-of-concept that across many different tasks we can substantially improve generalization.

\subsubsection{Bootstrapping with intermediate model sizes}
\label{sec:bootstrapping}

Bootstrapping is a long-standing idea in alignment: instead of directly aligning very superhuman models, we could first align an only slightly superhuman model, use that to align an even smarter model, and so on~\citep{christiano2019capability,
 christiano2018approval, superalignment, worley2021bootstrapped}. 
Our setting allows us to empirically test this idea.

\begin{figure}
    \centering
    \includegraphics[width=0.8\linewidth]{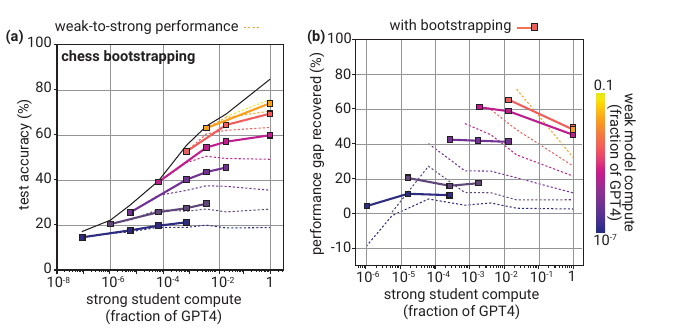}
    \centering
    \caption{
    \textbf{Bootstrapping improves weak-to-strong generalization on chess puzzles.}
    \textbf{(a)} Test accuracy as a function of strong student size. Accuracy of students trained with ground truth in black, accuracy of students naively trained with weak supervision shown with dotted lines (hue indicates size of weak supervisor). Accuracies of students trained via bootstrapping shown with colored squares (including both the final weak-to-strong performance and the performance of the intermediate models during bootstrapping).
    \textbf{(b)} Same as \textbf{a} with PGR.
    By taking multiple small steps instead of one big step we see substantially improved generalization, especially for larger student models.
    }
    \label{fig:chess-bootstrap}
\end{figure}

Specifically, we can construct a sequence of model sizes $\mathcal{M}_1 \rightarrow \mathcal{M}_2 \rightarrow \ldots \rightarrow \mathcal{M}_n$ of increasing sizes.
Then, we use the weak labels from $\mathcal{M}_1$ to finetune $\mathcal{M}_2$, use $\mathcal{M}_2$ to generate new weak labels that we can use to finetune the next model in the sequence, $\mathcal{M}_3$, and so on.

We evaluate bootstrapping in the chess puzzle setting. 
When we naively finetune on weak labels for chess (\Cref{section:baselines}), 
we see high PGR when we cross small supervisor-student gaps, but low PGR for larger gaps. 
As a result, in this setting it may help to take multiple small steps---steps where PGR should be high---instead of one big step.

For each round of bootstrapping, we run three iterations of weak-to-strong learning, i.e. we bootstrap the weak supervision using two intermediate model sizes before finally finetuning the largest model in the sequence.
We report the results (including all intermediate weak-to-strong models within each bootstrap) in \Cref{fig:chess-bootstrap}. 
Bootstrapping improves PGR compared to the baseline, especially for larger student models. With the naive method, transfer accuracy curves flatten as the weak-strong gap grows larger; with bootstrapping, the accuracy continues to monotonically improve.

While the results in the chess setting are promising, in preliminary experiments we observed only small improvements with bootstrapping on NLP tasks and no improvements in the RM setting. 
This makes sense intuitively: unlike in the chess setting where naive PGR decreased with larger supervisor-student gaps, naive PGR increased or was rougly constant for larger supervisor-student gaps in the NLP and reward modeling settings.
Overall, these results suggest bootstrapping is a plausible avenue to investigate for improving weak-to-strong generalization and can be helpful in some settings, but that naive bootstrapping alone will not be enough to align models much smarter than their supervisors.

\subsubsection{An auxiliary confidence loss can dramatically improve generalization on NLP tasks}
\label{sec:auxloss}

\begin{figure}
    \centering
    \includegraphics[width=0.8\linewidth]{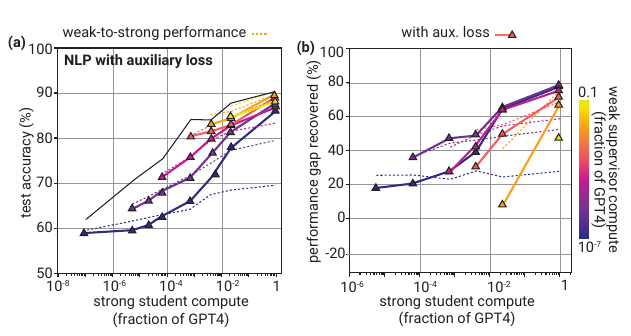}
    \caption{\textbf{Substantially improved generalization on NLP datasets with a simple auxiliary loss.}
\textbf{(a)} Test accuracy as a function of strong student size. Accuracy of a student trained with ground truth in black, accuracy of students naively trained with weak supervision shown with dotted lines. Accuracies of students trained with auxiliary confidence loss shown with colored triangles.  Median computed across 22 NLP tasks (hue indicates size of weak supervisor), see \Cref{fig:grid_all_nlp_transfer} for individual datasets. \textbf{(b)} Same as \textbf{a} with PGR.
The confidence loss can improve generalization drastically, especially for large supervisor-student gaps. 
    }
    \label{fig:nlp-generalization-method}
    \vspace{-10pt}
\end{figure}

In our baseline results (\Cref{section:baselines}), we naively finetune the strong student on the labels provided by the weak supervisor. 
Because we are directly training the strong student to imitate the weak supervisor, it may also learn to imitate the errors of the supervisor (see \Cref{sec:imitation} for more discussion).
Intuitively, we want to avoid this failure mode and provide additional regularization towards what the strong pretrained model already internally knows: we want the student to learn the intent of the supervisor, but not to imitate its mistakes.

We operationalize this intuition by adding an auxiliary confidence loss term to the standard cross entropy objective.
This method is closely related to conditional entropy minimization~\citep{grandvalet2004semi} which is a prominent technique in semi-supervised learning.
Specifically, we add an additional loss term which reinforces the strong model's confidence in its own predictions---even when they disagree with the weak labels.
We provide a detailed description of the method in \Cref{sec:app-aux-loss-details}.

In \Cref{fig:nlp-generalization-method}, we plot accuracy and PGR curves with this method on our NLP tasks. We find that while it performs slightly worse than the naive baseline for smaller strong students, it dramatically improves generalization for large gaps in compute between weak and strong models. With the smallest weak supervisor and largest strong student, the confidence loss increases median PGR from about 25\% to nearly 80\%.

\begin{figure}
    \centering
    \includegraphics[width=0.99\linewidth]{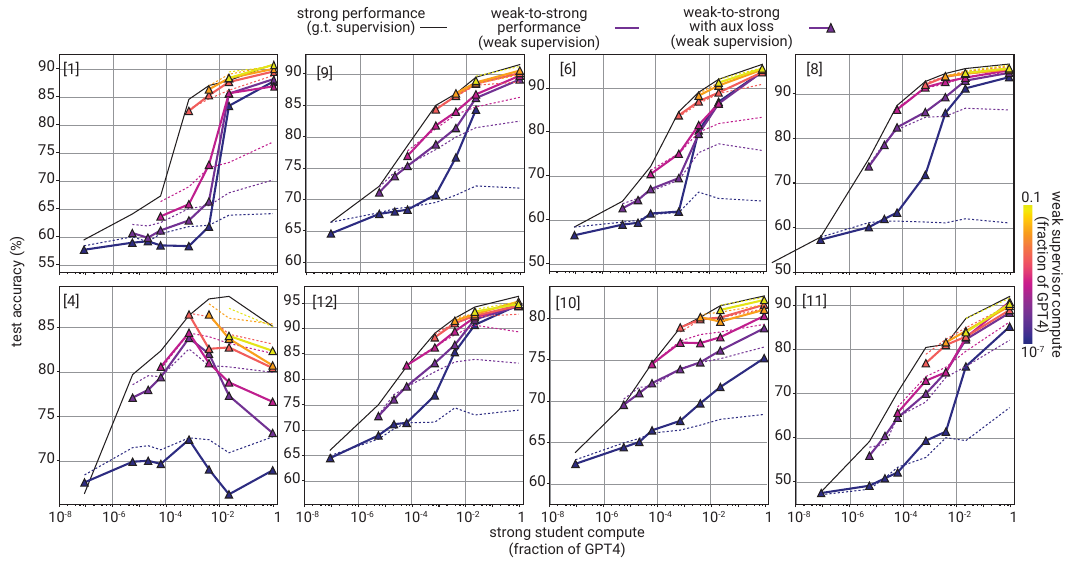} %
    \caption{\textbf{Simple auxiliary loss improves generalization across most datasets.} Test accuracy as a function of strong student compute for a representative sample of NLP tasks. See \Cref{table:datasets} for dataset details and Appendix Figure \ref{fig:all_grid_nlp} for results on all 22 NLP tasks.  Auxiliary loss is shown with triangles, and the baseline with dotted lines. Weak supervisor model size shown in varying colors, with ground truth supervision shown in black.
    }
    \label{fig:grid_all_nlp_transfer}
\end{figure}

In addition, we also plot generalization curves for a representative subset of NLP datasets in \Cref{fig:grid_all_nlp_transfer}, as well as the full panel of datasets in \Cref{fig:all_grid_nlp}.
There are some settings in which the confidence loss does not help much or degrades performance, e.g.~when the gap between the weak supervisor and strong student is small or when the dataset features inverse scaling even with ground truth supervision.
But the confidence loss improves performance on most NLP datasets dramatically, and for many datasets we get almost perfect generalization, recovering nearly all the performance of the strong model, even when using the smallest weak supervisors.

Finally, we find evidence consistent with our motivating intuition for the confidence loss (allowing the strong student to confidently disagree with its weak supervisor): the auxiliary loss reduces the strong student's imitation of weak errors and mitigates weak label overfitting (see \Cref{sec:imitation}).

\section{Understanding Weak-to-Strong Generalization}
\label{sec:understanding}

Strong methods will be essential for solving superalignment, but to trust those methods it is also important to understand \textit{when} and \textit{why} they work.
A better understanding of weak-to-strong generalization could help us trust that generalization will continue working even in the future high-stakes settings we care most about, and could help us develop better methods along the way.
In this section, we study two phenomena relevant to weak-to-strong generalization: imitation of supervisor mistakes and salience of the tasks to the strong student model.

\subsection{Understanding imitation}
\label{sec:imitation}

When we train a strong model with weak supervision on some task, our hope is that the strong model will perform that desired task as well as possible, leveraging the latent capabilities it learned from pretraining to significantly outperform the weak supervisor. 
A salient way in which we could fail to achieve that desired generalization is if the strong model instead learns to imitate the weak supervisor---predicting how the weak supervisor would have classified each example. In particular, if the weak labels contain systematic errors that are easy to learn, the strong model could learn to imitate those errors.
This is also a concern raised in theoretical work on superalignment, which has argued that the \emph{human simulator} failure mode could be important: naive human supervision might result in superhuman models learning to imitate what a human would say, rather outputting its best predictions~\citep{christiano2022eliciting}. 

\subsubsection{Overfitting to Weak Supervision}
\label{sec:results-overfitting}

\begin{figure}
    \centering
    \centerline{\includegraphics[width=1.0\linewidth]{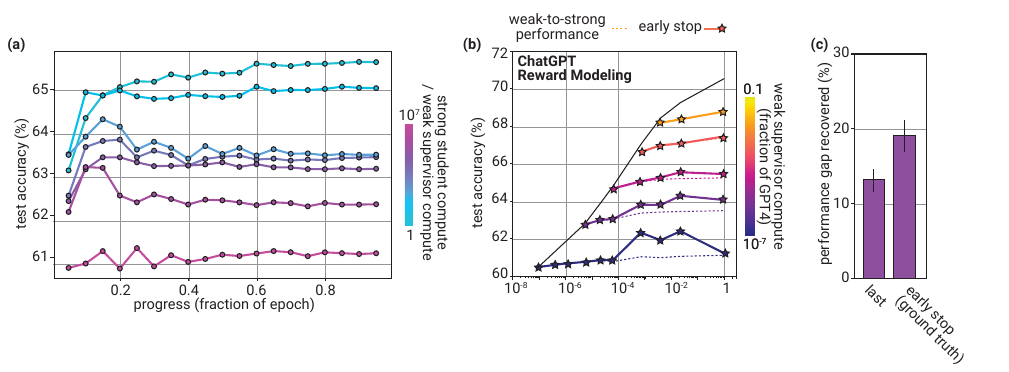}}
    
    \caption{\textbf{Strong models overfit to the weak labels.} 
    In all figures, we show data for the ChatGPT Reward Modeling task.
    \textbf{(a)} Weak-to-strong performance over the course of training. Hues indicate the student-supervisor gap.
    (\textbf{b}) Best weak-to-strong performance during training (stars) and weak-to-strong performance at the end of training (dashed). Weak performance in black. Hue indicates the size of the weak supervisor.
    (\textbf{c}) Median best and final performance gap recovered (PGR) aggregated across all supervisor-student pairs. %
    We see overfitting to weak labels for large weak-strong gaps, even within one epoch. In these cases, the best test accuracy achieved over training can be substantially better than the test accuracy at the end of training.
    See Figure \ref{fig:all_grid_nlp_overfit} for the corresponding analysis of a representative subset of NLP tasks.
    } 
    \label{fig:overfit}
\end{figure}

The failure mode of imitating weak supervision is especially relevant to our naive baseline in \Cref{section:baselines}, which directly trains the student to imitate the supervisor.
In the case of infinite training data, naively fitting the weak labels should result in perfect imitation, and a PGR of zero.  
In practice, we train on finite data for a small number of epochs.
Unlike typical ML settings, however, we could expect to observe overfitting even when training for less than a single epoch: the strong model might overfit to the weak supervisor labels and its errors, degrading ground truth test accuracy over training even without classic overfitting to any specific training examples.

Empirically, we see that the strong student indeed appears to overfit to the weak supervisor's errors. 
In \Cref{fig:overfit}(a) we show ground truth test accuracy curves over the course of training for the ChatGPT RM task, and in \Cref{fig:overfit}(b) and (c) we compare the best\footnote{Note that our best test accuracies may slightly overstate accuracy, due to noisy evaluations.} and final ground truth test accuracies (median across all weak-strong model pairs). 
We find overfitting for large weak-strong gaps. For small weak-strong gaps, weak-to-strong performance typically monotonically increases over the course of training. For larger gaps, weak-to-strong performance often increases initially, but then starts dropping well before a single epoch has elapsed.  
Ground truth early stopping, which ``cheats'' by evaluating against ground truth and stopping at an optimal step with respect to ground truth test labels, typically gives a PGR improvement of around $5$ percentage points.

We see the same phenomenon for NLP tasks in \Cref{fig:all_grid_nlp_overfit}. 
In the NLP setting, we find that ``cheating'' early stopping on ground truth gives a $15$ percentage point boost in PGR over the model at the end of training, and a $10$ percentage point boost in PGR compared to ``non-cheating'' early stopping with respect to weak labels.

Unfortunately, an early stopping criterion that uses ground truth labels does not constitute a valid method. Nevertheless, the results above suggest that imitating weak supervisor errors may be an important phenomenon in our setting.

Moreover, these results suggest that better early stopping or regularization strategies may be able to substantially improve weak-to-strong generalization, by reducing overfitting to the weak labels and their errors.
Indeed, we see in \Cref{fig:all_grid_nlp_overfit} that the auxiliary confidence loss introduced in \Cref{sec:auxloss} reduces overfitting to weak labels on NLP tasks substantially. For large weak-strong gaps, early stopping on ground truth (compared to early stopping on weak labels) gives a 15\% PGR boost when using the naive method, but only a roughly 5\% PGR boost when using the confidence loss.

\subsubsection{Student-supervisor agreement}

Another way to measure imitation is to directly measure the agreement between the student and the supervisor: the fraction of test inputs where the strong student makes the same prediction as the weak supervisor.  Note that if agreement were 100\%, then weak-to-strong accuracy would be equal to supervisor accuracy, and PGR would be 0.  

In general, we notice that for our naive finetuning baseline, student-supervisor agreement is consistently high---often noticeably higher than weak supervisor accuracy.
This indicates that the student is imitating some of the supervisor's errors. 
These phenomena hold across all tasks (NLP tasks, chess, and reward modeling) and all model sizes, for the naive method.

The confidence loss in \Cref{sec:auxloss} reduces student-supervisor agreements significantly (\Cref{fig:agreement_breakdown}), primarily by imitating supervisor mistakes less (\Cref{fig:agreement_breakdown}c).
The loss encourages the strong student to make confident predictions, including when they contradict the weak supervisor.
In a handful of the settings where it is most successful, the confidence loss reduces student-supervisor agreement below strong student test accuracy (weak-to-strong performance)---i.e., the resulting model is fitting the ground truth concept \textit{better} than it is fitting the weak labels it was trained with.

\subsubsection{Inverse scaling for imitating the supervisor}
\label{subsec:emulating-weak}

\begin{figure}
    \centering
    \includegraphics[width=0.95\linewidth]{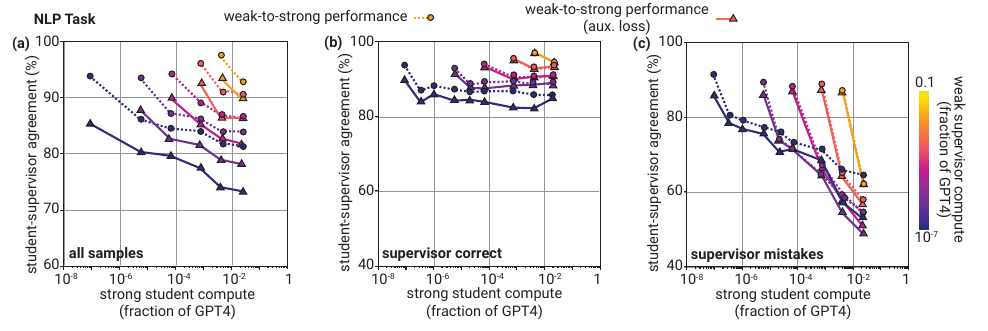}
    \caption{
    \textbf{Student-supervisor agreement \textit{decreases} with larger student-supervisor gaps; the confidence loss reduces imitation of supervisor mistakes.}
    \textbf{(a)} Student-supervisor agreement as a function of strong student size on NLP tasks, \textbf{(b)} \textbf{a} but only on samples where the supervisor is correct,  \textbf{(c)} \textbf{a} but only on samples where the supervisor is mistaken.
    Dotted lines indicate naive finetuning on weak labels, and triangles indicate results with the auxiliary confidence loss results (see \Cref{sec:method}).
    Hue of line indicates size of weak supervisor.
 For results on reward models, see Figure \ref{fig:agreement_breakdown_rm}.
    } 
    \label{fig:agreement_breakdown}
    \vspace{-10pt}
\end{figure}

Next, we study student-supervisor agreement as a function strong model size (see \Cref{fig:agreement_breakdown} and \Cref{fig:agreement_breakdown_rm}). Surprisingly, we find inverse scaling~\citep{mckenzie2023inverse}: larger student models consistently agree \emph{less} with the errors of the supervisor than smaller student models, despite being trained to imitate the supervisor, not using early stopping, and having larger capacity than smaller student models. 

This trend is especially strong if we evaluate agreement only on datapoints where the supervisor is wrong (\Cref{fig:agreement_breakdown}c), and the trend persists if looking at cross entropy loss instead of accuracy.  

These results suggest that pretrained models may have a hard time fitting errors of other (smaller) pretrained models, at least in finetuning settings with relatively limited data. 
\citet{stanton2021does} and \citet{furlanello2018born} report a related observation in the context of knowledge distillation: it is surprisingly hard for models to fit the predictions of other models, even when they have sufficient capacity to do so.

One natural hypothesis is that the nature of (especially naive) weak-to-strong generalization depends heavily on the error structure of the weak supervisors and how easy those errors are to imitate.  
In \Cref{sec:app-simulation}, we show initial experiments that test how different types of weak supervision errors impact what the strong student learns. Our results suggest that errors that are more difficult for the student to imitate result in stronger naive weak-to-strong generalization, but that even when they are easy to imitate, the confidence loss can help.

\subsection{Saliency in the strong model representations}
\label{sec:salience}

One intuition for when weak-to-strong generalization might be feasible is when the task or concept we want to elicit is internally ``salient'' to the strong model.
In this section, we study several phenomena related to the saliency of the concepts we are trying to elicit from the student model.

\subsubsection{Eliciting strong model knowledge with prompting}
\label{subsec:prompting}

\begin{figure}
    \centering
    \includegraphics[width=0.95\linewidth]{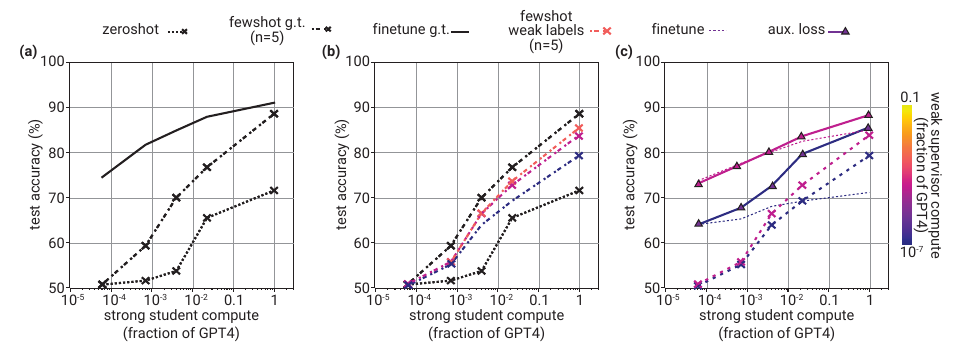}
    \caption{\textbf{Few-shot prompting becomes competitive with finetuning for large models; weak-to-strong learning is qualitatively similar in the prompting setting.}
    \textbf{(a)} Average zero-shot (single dashed), 5-shot (double dashed) and finetuning (solid) accuracy with ground truth labels as a function of strong student size.
    \textbf{(b)} Average 5-shot with weak labels (colored dashed) accuracy as a function of student model size. Hue of line indicates size of weak supervisor. Zero-shot and 5-shot same as in panel \textbf{a}.
    \textbf{(c)} Average weak-to-strong performance for 5-shot prompting (dashed with crosses), naive finetuning (dashed thin) and finetuning with the confidence loss (solid with triangle) as a function of student model compute.
    Results are averaged across $7$ NLP tasks.
    Few-shot weak-to-strong performance becomes competitive with or outperforms finetuning for the largest strong students, though finetuning with the confidence loss does better.
    }
    \label{fig:fewshot}
    \vspace{-10pt}
\end{figure}

One possible reason for the high PGR we observe in \Cref{sec:results} could be that eliciting what the strong model knows is easy. 
In particular, it is possible that strong pretrained models can solve many relevant tasks zero-shot with a simple prompt. 

In \Cref{fig:fewshot}a, we consider $7$ representative NLP tasks and compare finetuning, zero-shot prompting, and 5-shot prompting; for this initial experiment, we use ground truth labels rather than weak labels for finetuning and 5-shot.
For both the zero-shot and 5-shot baseline we use task-specific prompts summarized in \Cref{tab:prompts}.
We find that zero-shot and 5-shot test accuracy is poor for most model sizes but, consistent with \cite{brown2020language}, improves drastically for larger model sizes.
In particular, for the largest models, 5-shot prompting becomes competitive with finetuning on many tasks, indicating that eliciting the task-relevant knowledge of these very large models is relatively straightforward.

We are also interested in weak-to-strong learning in the context of few-shot prompting.
To study this setting, we construct a few-shot prompt where the labels are provided by the weak supervisor.
We report the results in \Cref{fig:fewshot}b.
Consistent with our findings in the finetuning setting, we get worse performance when we few-shot prompt with weak labels than we do few-shot prompting with ground truth labels. This suggests that weak-to-strong learning is a nontrivial problem in the prompting setting as well.

Similar to the finetuning setting, few-shot weak-to-strong performance improves for stronger supervisors.
Compared to our weak-to-strong finetuning baseline (\Cref{fig:fewshot}c), weak-to-strong performance of few-shot prompting is poor for smaller student models, but becomes competitive or even outperforms finetuning for the largest strong students.  
However, weak-to-strong finetuning with the confidence loss still generally outperforms weak-to-strong few-shot prompting.

Overall, these results provide an important reference for our results on weak-to-strong generalization.
They suggest that for the largest model sizes, the knowledge needed to solve many task can be elicited fairly easily with prompting.
However, our current setup may be more disanalogous for prompting than for finetuning; many of our NLP tasks may have been implicitly observed during pretraining, which we conjecture benefits prompting more than finetuning. 
We discuss this potential disanalogy much more in \Cref{subsec:disanalogies}.

\subsubsection{Generative supervision improves RM weak-to-strong generalization}
\label{sec:rm-sft}

\begin{figure}
    \centering
    \includegraphics[width=0.8\linewidth]{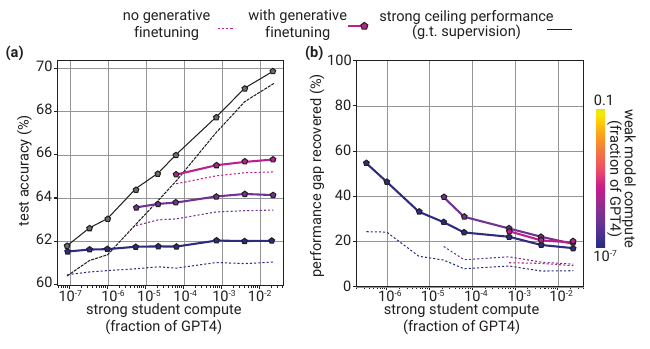}
    \caption{
    \textbf{Generative finetuning on reward modeling data improves weak-to-strong performance and PGR.}
    \textbf{(a)} Weak-to-strong performance on the reward modeling task, with (solid lines) and without (dashed lines) an extra step of generative finetuning for the strong student model.  
    Solid black line shows a strong ceiling reward model that was also trained with the generative finetuning step; dashed black line show a weak supervisor reward model trained \textit{without} the generative finetuning step. 
    \textbf{(b)} PGR with and without generative finetuning. For generative finetuning PGR, we use the strong ceiling performance that also had this extra generative finetuning step. Even with this ceiling adjustment, PGR is higher with an extra generative finetuning step.
    }
    \label{fig:rm_sft}
    \vspace{-10pt}
\end{figure}

If salient representations of the desired task is useful for weak-to-strong generalization, then we may be able to improve generalization by increasing the salience of the task to the strong model.
One way to increase the salience of a task without needing ground truth labels is to perform unsupervised finetuning with the language modeling objective on data relevant to that task~\citep{dai2015semi}.
For example, by finetuning a language model in an unsupervised way on online reviews, sentiment becomes saliently represented to models internally~\citep{radford2017learning}.

We test this idea in our reward modeling setting, where it is standard practice to initialize the model with a baseline finetuned on demonstrations of desired behaviors~\citep{stiennon2020learning}. 
In our case, we re-use the ChatGPT comparison data instead of introducing a new supervision dataset. 
Comparisons are comprised of a prefix (a single request or conversation between the user and assistant) and at least two candidate completions.
We finetune the base models with a language modeling loss on all prefix-completion pairs, ignoring the human preferences between those completions.

Note that these pairs include completions ranked worst by human raters, so this procedure should not in principle leak any information about the ground truth preference labels that the weak-to-strong models should not have access to. On the other hand, since the completions can come from humans or stronger models, there may be some leakage similar in kind to the pretraining leakage that we discuss as a disanalogy in \Cref{subsec:disanalogies}.  
Even in this setup, the reward modeling task is highly non-trivial, and we leave addressing this disanalogy (e.g.~by collecting completions only from weaker models) for future work.

We found that the additional generative finetuning on the RM data leads to better weak-to-strong performance. 
Because this procedure also improves the performance of models trained on ground truth RM data, we compare our new weak-to-strong performance to strong ``ceiling'' models that were also first generatively finetuned in the same way.
Even with this adjusted ceiling, we find that generative supervision improves PGR by approximately 10-20\%.  We report the results in \Cref{fig:rm_sft}.

Furthermore, the improvement from generative finetuning stacks with the improvement from ground truth early-stopping (a ``cheating'' method to illustrate potential performance if we could optimally early stop, see \Cref{sec:results-overfitting}).
When we combine these two techniques, we can achieve PGR of approximately 30-40\%, which would make the results on the RM task competitive with the weak-to-strong generalization we observe on NLP and chess puzzle tasks.

We can apply the idea of improving task saliency with generative finetuning on relevant data to all settings, and we believe this could be a promising direction for future work.

\subsubsection{Finetuning on weak supervision to increase concept saliency}

One possible measure of concept saliency is how linearly represented a task is.  In particular, we can measure the performance of a linear probe (logistic regression classifier) trained from frozen activations of the model.
If the optimal solution can be approximately recovered with a linear probe, that could simplify our problem greatly; we could focus on linear probing methods instead of finetuning methods, which could greatly reduce the search space we need to consider to elicit the desired generalization.  In our work, we focus only on how linearly represented a task is in the final activations, prior to the unembedding layer.

\begin{figure}
    \centering
    \includegraphics[width=0.5\linewidth]{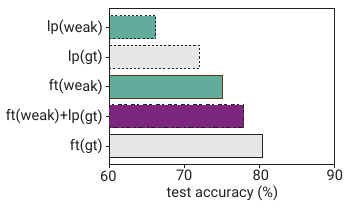}
    \caption{\textbf{Finetuning on weak supervisor labels makes the desired generalization more linearly represented.} We plot test accuracy for five different strategies, averaged across a subset of NLP tasks. \textbf{lp(weak)}: training a linear probe on the base model using weak labels, \textbf{lp(gt)}: training a linear probe on the base models using ground truth labels, \textbf{ft(weak)}: finetuning the model on weak labels, \textbf{ft(weak) + lp(gt)}: finetuning the model on weak labels \textit{then} training a linear probe on ground truth labels, \textbf{ft(gt)}: finetuning the model on ground truth labels.
    Finetuning on the weak labels significantly increases the linearity of the ground truth concept.
    }
    \label{fig:ftlp}
    \vspace{-10pt}
\end{figure}

In \Cref{fig:ftlp}, we plot average test accuracy on a subset of our NLP datasets for several different combinations of (1) finetuning or linear probing, using (2) weak or ground truth labels. 
First, we show linear probes trained with ground truth labels (72\% accuracy on average) perform worse than finetuning with ground truth labels (82\% on average), indicating that the optimal solution to most tasks is \textit{not} represented completely linearly in the strong model's final activations.
For comparison, we also report the results for linear probing and finetuning using weak labels, which we verify are worse than using ground-truth labels.

However, we find that we can achieve substantially better performance by \textit{first} finetuning the model on the \textit{weak} labels, and \textit{then} linear probing using the \textit{ground truth} labels.
In other words, when we finetune the strong model with weak labels, the representations become \textit{more linear even with respect to ground truth labels}.
In fact, finetuning on weak labels then linear probing on ground truth labels results in an accuracy of 78\%, closing 60\% of the gap between ground truth linear probing and finetuning.
This also noticeably outperforms the naive weak-to-strong finetuning baseline.

This phenomenon is closely related to a recent finding reported by \citet{kirichenko2023last} in the spurious cues literature.
They find that finetuning a model on biased supervision can result in models with very biased outputs, but surprisingly strong linear representations of the desired concepts.
These results suggest an alternative approach to improving weak-to-strong generalization. We could first ``linearize'' the desired concept, e.g.~by naively finetuning on weak labels. Then we could use simpler linear probe-based weak-to-strong methods to elicit the desired concept.

\section{Discussion}
\label{sec:discussion}

In this paper, we proposed a simple analogy for studying a core challenge of aligning superhuman models and showed that it is feasible to make significant progress on this problem.
However, our setup still has important disanalogies, which we now elaborate on.
We then outline a number of promising avenues for future work.

\subsection{Remaining disanalogies}
\label{subsec:disanalogies}

\textbf{Imitation saliency: superhuman models may easily imitate weak errors.}\quad
Future models will likely be very good at predicting what humans will think and say, especially if they are trained on human data in a similar manner to current models. Consequently, if we naively train such a superhuman model with human supervision, it might simply imitate the weak supervisor, outputting human-level capabilities rather than its latent superhuman capabilities ~\citep{christiano2022eliciting}.

This problem is only partially captured by our setup. While our strong pretrained models do imitate weak supervisors to some extent, they are not explicitly pretrained to imitate weak models, and our results from \Cref{subsec:emulating-weak} suggest that larger strong models may even have more difficulty doing this imitation. As such, ``imitating the weak supervisor'' may not be as much of a problem in our setup as it will be for the ultimate superalignment problem.  This may inflate generalization performance today.
We believe a more thorough investigation of this problem is an important area for future work.

\textbf{Pretraining leakage: superhuman knowledge may be latent, not observable.}\quad
\label{sec:latent_vs_observable}
Many of the tasks we consider in this work may have been observed in pretraining at least indirectly, for example through questions on online forums or through slight reframings of the task. 
For example, it is highly likely that simple science questions similar to those in the SciQ NLP task are present in our GPT-4 series pretraining dataset at least implicitly in some form. 
However future superhuman models may never directly observe superhuman alignment-relevant capabilities; these capabilities may be predominantly ``latent'', e.g.~learned through self-supervised learning or reinforcement learning rather than through imitation learning.
Intuitively, latent capabilities may be harder to elicit than capabilities that models could have observed in their pretraining data.

This disanalogy could cause our results to be overly optimistic. 
We conjecture that this disanalogy also increases prompting performance (\Cref{subsec:prompting}) more than it increases finetuning performance; intuitively prompting may work especially well on tasks that the model assigns high probability to observing.
If so, this would make prompting more disanalogous in our setup than finetuning.
We hope to test this conjecture in future work.

In \Cref{app:sec:imagenet}, we show a proof of concept that weak-to-strong generalization can still elicit latent capabilities that were never explicitly observed during pretraining, and even when prompting is not possible. 
In particular, we use AlexNet~\citep{krizhevsky2012imagenet} to supervise models pretrained with DINO~\citep{caron2021emerging}, a self-supervised method in computer vision that learns strong representations.
We find that the strong student generalizes significantly beyond AlexNet's performance, even though the student never observed any classification labels during pretraining.
Future work should study and mitigate this pretraining leakage disanology more systematically.

\subsection{Future Work}
\label{sec:future-work}

What would convince us that we have a ``solution'' to superalignment? 
This is a complicated question and we do not claim to have a complete answer. 
However, we expect substantial progress in at least the following three areas will be necessary: analogous setups, scalable methods, and strong scientific understanding. We now sketch out concrete problems for each of these areas.

\subsubsection{Concrete Problems: Analogous Setups}

Having strong measurements and a reliable methodology is extremely important for making empirical progress in any field. 
In particular, it is important that we have metrics which provide strong signal about whether we are making real progress toward the problem we ultimately care about.
Important directions for follow-up work include:

\begin{itemize}
    \item Making our setup more analogous by fixing the main remaining disanalogies described in \Cref{subsec:disanalogies}. Analogous setups are essential to ensure that methods that work today will continue to work for superhuman models.
    \item Validating that disanalogies are not severe, for example by checking that results are qualitatively similar to using e.g.~3rd grade humans to supervise our strongest models today.
    \item Relaxing some of the simplifications we made, e.g.~by generalizing our methods and results to complicated generative tasks.
    \item Testing how robust our weak-to-strong classifiers are to optimization pressure when we attain high PGR; for example, if we attain good weak-to-strong generalization with RMs, can we optimize the learned RM using RL?
    \item Testing our conjecture that prompting-based methods in our current setup will not be as indicative of future results relative to finetuning-based methods (\Cref{subsec:prompting}), and improving our setup to fix this.
    \item Identifying new or more specific disanalogies with our setup and fixing them.
\end{itemize}

Additionally, we do not yet know what future models will look like. We should update our setup over time as we learn more about how broadly superhuman models will be built.

\subsubsection{Concrete Problems: Scalable Methods}\label{subsec:concrete_problems_scalable_methods}

One intuition for why major progress on weak-to-strong generalization seems possible is because all we need to do is extract everything the strong model already ``knows'' about the task of interest---the strong model should intuitively already understand the task, and should hopefully have salient representations of that task.
This suggests a number of properties that should be satisfied by the desired generalization, and which we may be able to measure without access to ground truth.
\begin{itemize}
    \item The desired generalization should be able to \emph{disagree with the weak supervision} when the weak supervision is wrong. This is a property our auxiliary confidence loss may capture.
    \item The desired generalization should be ``\emph{natural}'' or ``\emph{salient}'' to the model. For example, we should not need to change the model too much to elicit the desired concept.
    \item The desired generalization should be \emph{consistent}. Consistency properties range anywhere from basic logical consistency to complicated forms of consistency between many prompts (e.g.~cycle consistency, cross examination, etc.).
\end{itemize}

Future work should identify additional unsupervised properties that can be used to specify the desired generalization.
More generally, there are very likely existing methods in the machine learning literature (e.g.~in semi-supervised learning or robust finetuning), which would be natural to try and which could also lead to substantial gains in weak-to-strong generalization. 
Generalization-based approaches to weak-to-strong learning are complementary to scalable oversight methods, in which the weak supervisor interacts with the strong model to improve the quality of the weak supervision.

\subsubsection{Concrete Problems: Scientific Understanding}

We will need an extremely high degree of trust and reliability in our methods for aligning superhuman models in high-stakes settings. We will not get this from strong benchmark performance alone. Instead, we also need a thorough understanding of precisely \textit{when} and \textit{why} our methods work. Example questions of interest include:

\begin{itemize}
    \item What explains the difference between the relatively strong results on NLP datasets and the relatively poor results with reward models when using naive finetuning? 
    \item What makes a concept easy or hard to elicit?  What is a good definition of ``salience''?
    \item Can we reliably estimate generalization error at test time without any labels? For example, can we measure the degree of weak-to-strong underspecification~\citep{lee2022diversify}? 
    \item Can we reliably extrapolate generalization error across many orders of magnitude using scaling laws?
    \item How important are the errors in the weak supervision, precisely? How do different kinds of weak label biases affect generalization?  
    \item How robust are our proposed methods to optimization pressure? 
\end{itemize}

In \Cref{sec:understanding} we only scratched the surface for understanding weak-to-strong generalization, but future work will need to go much further. 
An advantage of our setup is that it makes it easy to run simple experiments to scientifically study generalization phenomena across a wide range of settings.

\subsection{Conclusion}

Recent progress in AI has been faster than almost anyone anticipated~\citep{steinhardt2022forecasting,bengio2023managing}. 
For an increasing number of researchers, the possibility of superhuman models being developed this decade has become increasingly plausible.
Broadly superhuman models would be extraordinarily powerful and, if misused or misaligned with humans values, could potentially cause catastrophic harm~\citep{ai-risk-open-letter}. %
Given the stakes, we need to establish extremely high reliability in the alignment of these systems ahead of time.
But for years it has been unclear how to empirically study superhuman model alignment. 
We believe it is now easier to make progress on this problem than ever before.

\section{Acknowledgements}
We would like to thank Boaz Barak, Paul Christiano, Jacob Steinhardt, Ananya Kumar, Jakub Pachocki, John Schulman, Wojciech Zaremba, Alec Radford, Nat McAleese, and William Saunders for valuable technical insights and discussions.  
We are grateful to Mia Glaese, Boaz Barak, Kush Bhatia, Jean-Stanislas Denain, Erik Jones, Polina Kirichenko, Daniel Kokotajlo, Yoonho Lee, Jessy Lin, Richard Ngo, John Schulman, Peter Tong, Fred Zhang, Ruiqi Zhong, Ryan Greenblatt, Fabien Roger, Paul Christiano, Steven Adler, Rai Pokorny, Adam Kalai, Jacob Hilton, Roger Grosse, Dan Hendrycks, Alec Radford, and Scott Aaronson for helpful feedback on earlier drafts of this paper.
We also thank Shantanu Jain, Avital Oliver, Suchir Balaji, Cathy Yeh, and the Platform team for infrastructure help.
CB is also grateful to Dan Hendrycks, Jacob Steinhardt, and Paul Christiano for many formative discussions over the years.

\bibliography{references}

\newpage
\appendix
\section*{Appendix Outline}
\begin{itemize}
    \item In \Cref{app:sec:details}, we provide additional details on our setup and experiments.
    \item In \Cref{sec:app-other-methods}, we describe additional results, including negative results and methods that did not work well in our experiments.
    \item In \Cref{sec:app-easy-hard}, we report results on easy-to-hard generalization, where we only provide supervision on easy examples.
    \item In \Cref{sec:app:more-weak-to-strong}, we provide results in two more weak-to-strong learning settings: 
 a self-supervised computer vision setting on ImageNet, and a pure linear probing setting.
    \item In \Cref{sec:app-simulation}, we provide additional results and discussion on the effect of weak supervisor error simulation.
    \item In \Cref{sec:app-how-study}, we discuss how we believe methodological progress should be made on superalignment.
    \item In \Cref{sec:app-alignment}, we describe how our work fits into the bigger picture of alignment.
\end{itemize}

\section{Further experimental details}
\label{app:sec:details}

Here, we provide further details on our experiments.
Across all tasks, we use pretrained base models from the GPT-4 family~\citep{openai2023gpt}, spanning a range of model sizes. 

\subsection{NLP Tasks}

\begin{figure}
    \centering
    \includegraphics[width=1\linewidth]{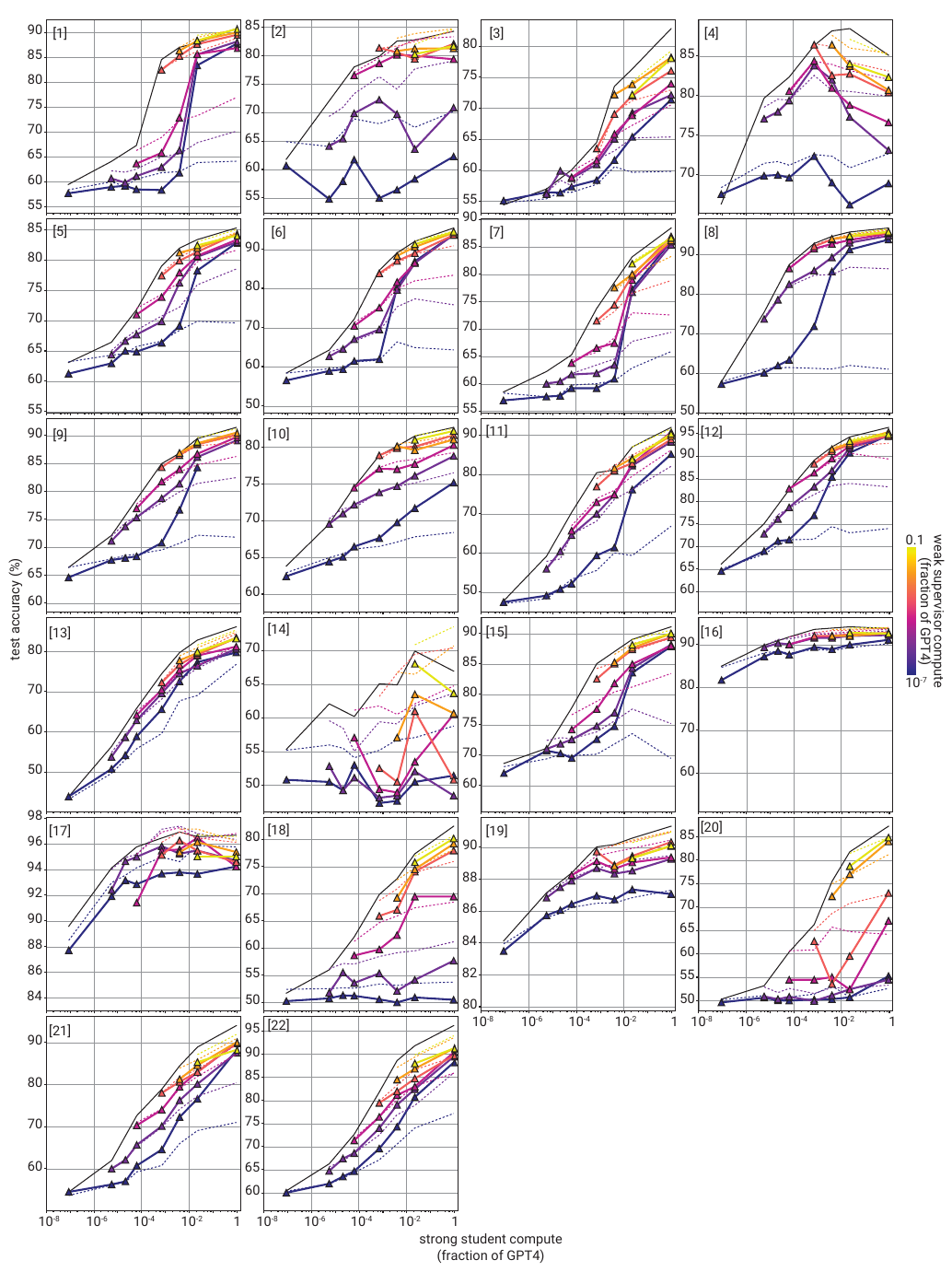}
 \caption{\textbf{Full weak-to-strong generalization results across 22 NLP datasets.} Test accuracy as a function of strong student compute across our full suite of standard NLP tasks. See \Cref{table:datasets} for dataset details.}
    \label{fig:all_grid_nlp}
\end{figure}

\textbf{Data preprocessing.}\quad
We use popular NLP classification benchmark datasets listed in \Cref{table:datasets}.
We obfuscate the names of the datasets in our plots (e.g.~\Cref{fig:all_grid_nlp}) for confidentiality;
across all figures, we replace the names of the datasets with their order in a randomized sequence.
We apply various preprocessing to the datasets. For example, some tasks are in FLAN~\citep{weifinetuned} and we use their preprocessing.  For ANLI we group neutral entailments with contradictions.
We convert each dataset to a binary classification problem.
For multiple-choice datasets, suppose each datapoint has a question $Q$ and multiple candidate answers $A_1, \ldots, A_k$.
We then convert this datapoint to $k$ new datapoints of the form $(Q, A_i)$, where the label is $0$ for all incorrect answers $A_i$ and $1$ for the correct answers.
In this procedure, we also aim to maintain class balance, so we keep the same number of correct and wrong answers per question\footnote{In some datasets there are multiple correct answers for each question.}.
We are also additionally rebalancing the classes in datasets where one of the classes represents more than $55\%$ of the data.
To do so, we randomly drop datapoints from the dominant class, so that the classes are perfectly balanced.

\textbf{Models.}\quad In order to adapt our language models to the classification setting, we replace the unembedding layer of the model with a linear classification head with two outputs.
We initialize the weights of the classification head with the unembedding weights for tokens ``0'' and ``1''.

\textbf{Training hyperparameters.}\quad We finetune all models for 2 epochs using a batch size of 32. 
In the weak-to-strong generalization experiments, we early stop training based on the accuracy with respect to the weak labels on a held-out validation set.
See \Cref{sec:results-overfitting} for relevant discussion.
We only tuned the hyper-parameters of our methods on smaller model sizes, and on a subset of 8 datasets.
The full GPT-4 model and most of the datasets were held-out, except for datasets [5--12] (see \Cref{fig:all_grid_nlp}).

\textbf{Weak labels.} \quad
To produce the weak labels, we split the original dataset in half. 
We ensure that related datapoints, e.g.~datapoints that share the same question or premise, are always grouped together into the same half.
Then, we train the weak supervisor model on the first half of the dataset, and use its prediction on the other half as the weak labels.
We additionally save the weak labels on the test set to evaluate metrics such as agreement in \Cref{subsec:emulating-weak}.
The weak labels are soft labels on the training data, i.e. the class probabilities predicted by the supervisor. 

\textbf{Evaluation.}\quad
For all datasets, we report accuracy on the test set which is also balanced to have an equal number of datapoints in each class.
In particular, random guess performance corresponds to $50\%$ accuracy on all NLP datasets.

\begin{table}[t]
\centering
\caption{\textbf{Datasets and their sources.}  \label{table:datasets}
We summarize the NLP datasets we use and their original sources.}
\begin{tabular}{ll}
\toprule
Dataset & Original Source \\
\midrule \midrule
\href{https://arxiv.org/abs/1905.10044}{BoolQ} & \citet{clark2019boolq} \\
\href{https://www.aclweb.org/anthology/D19-1243}{CosmosQA} & \citet{huang2019cosmos} \\
\href{https://direct.mit.edu/tacl/article/doi/10.1162/tacl_a_00264/43501/DREAM-A-Challenge-Data-Set-and-Models-for-Dialogue}{DREAM} & \citet{sun2019dream} \\
\href{https://arxiv.org/abs/2008.02275}{ETHICS [Justice]} & \citet{hendrycks2020aligning} \\
\href{https://arxiv.org/abs/2008.02275}{ETHICS [Deontology]} & \citet{hendrycks2020aligning} \\
\href{https://arxiv.org/abs/2008.02275}{ETHICS [Virtue]} & \citet{hendrycks2020aligning} \\
\href{https://arxiv.org/abs/2008.02275}{ETHICS [Utilitarianism]} & \citet{hendrycks2020aligning} \\
\href{https://arxiv.org/abs/2109.01652}{FLAN ANLI R2} & \citet{nie2019adversarial,weifinetuned} \\
\href{https://gluebenchmark.com/}{GLUE CoLA} & \citet{warstadt2019neural}; \citet{wang2018glue} \\
\href{https://gluebenchmark.com/}{GLUE SST-2} & \citet{socher2013recursive}; \citet{wang2018glue} \\
\href{https://arxiv.org/abs/1905.07830v1}{HellaSwag} & \citet{zellers2019hellaswag} \\
\href{https://aclanthology.org/D19-1332/}{MCTACO} & \citet{ZKNR19} \\
\href{https://arxiv.org/abs/1809.02789}{OpenBookQA} & \citet{OpenBookQA2018} \\
\href{https://arxiv.org/abs/1904.01130}{PAWS} & \citet{paws2019naacl} \\
\href{https://ojs.aaai.org/index.php/AAAI/article/view/6398}{QuAIL} & \citet{rogers2020getting} \\
\href{https://arxiv.org/abs/1911.11641}{PIQA} & \citet{Bisk2020}\\
\href{https://arxiv.org/abs/1909.03553}{QuaRTz} & \citet{quartz} \\
\href{https://arxiv.org/abs/1707.06209}{SciQ} & \citet{welbl2017crowdsourcing} \\
\href{https://arxiv.org/abs/1904.09728}{Social IQa} & \citet{sap2019socialiqa} \\
\href{https://super.gluebenchmark.com/}{SuperGLUE MultiRC} & \citet{khashabi2018looking}; \citet{wang2019superglue} \\
\href{https://super.gluebenchmark.com/}{SuperGLUE WIC} & \citet{pilehvar2018wic}; \citet{wang2019superglue} \\
\href{http://thinknook.com/twitter-sentiment-analysis-training-corpus-dataset-2012-09-22/}{Twitter Sentiment} & \citet{paws2019naacl} \\
\bottomrule
\end{tabular}
\end{table}

\begin{figure}
    \centering
    \includegraphics[width=1\linewidth]{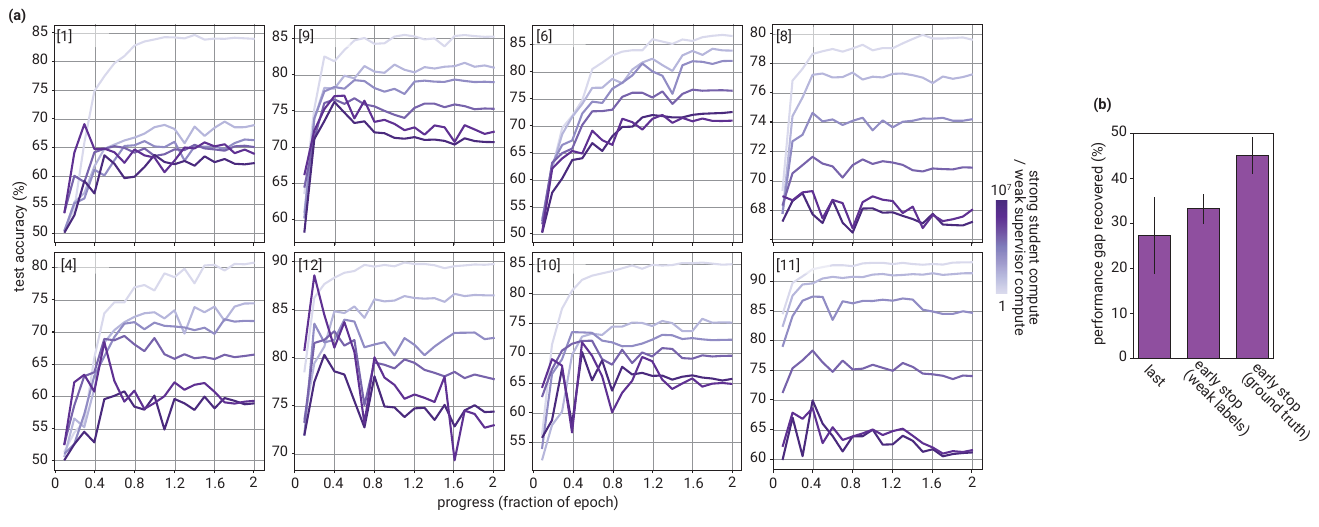}
 \caption{\textbf{Overfitting during training, for NLP datasets.} \textbf{Strong models overfit to the weak labels.} 
    \textbf{(a)} Ground truth test accuracy of strong students over the course of training for a subset of our NLP task. Hues indicate the gap between weak supervisor and strong student model compute. Inset numbers indicate dataset id (compare \Cref{fig:all_grid_nlp}).
    (\textbf{b}) Median best, early-stopped according to weak label agreement, and final performance gap recovered (PGR) aggregated across all supervisor-student pairs and all NLP tasks. Error bars indicate standard error of the mean (s.e.m.).}
    \label{fig:all_grid_nlp_overfit}
\end{figure}

\paragraph{Detailed results.} In \Cref{fig:all_grid_nlp}, we provide detailed results across all datasets for both the baseline and the auxiliary confidence loss introduced in \Cref{sec:method}.
In \Cref{fig:all_grid_nlp_overfit} we report the detailed results on overfitting to the weak supervisor predictions for the NLP datasets.

\subsection{Chess Puzzles}
\label{sec:app-details-chess}

\begin{figure}
    \centering
    \begin{tabular}{cc}
        \hspace{-0.5cm}\includegraphics[width=0.3\linewidth]{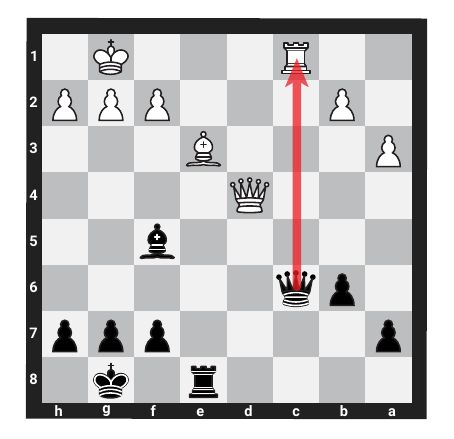} &
        \hspace{-0.5cm}\includegraphics[width=0.3\linewidth]{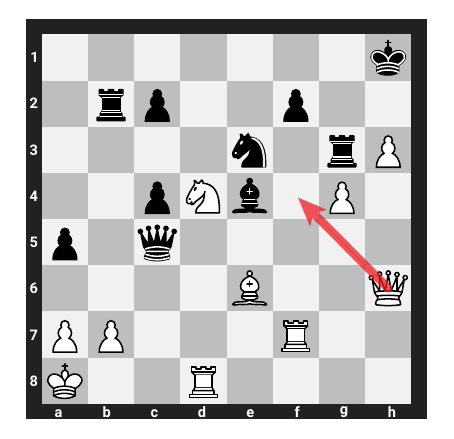} \\[0.2cm]
        \hspace{-0.5cm}
        \begin{tabular}{l}
        \tiny{\textbf{Prompt:}  ``1. d4  1... Nf6  2. Nf3  2... d5  3. e3  3... e6  4. Bd3  4... c5} \\[-0.15cm]
        \tiny{5. c3  5... Be7  6. Nbd2  6... O-O  7. O-O  7... Nc6  8. Re1  8... Bd7  9. e4  9... dxe4} \\[-0.15cm]
        \tiny{10. Nxe4  10... cxd4  11. Nxf6+  11... Bxf6  12. cxd4  12... Nb4  13. Be4  13... Qb6} \\[-0.15cm]
        \tiny{14. a3  14... Nc6  15. d5  15... exd5  16. Bxd5  16... Bf5  17. Bxc6  17... Qxc6} \\[-0.15cm]
        \tiny{18. Nd4  18... Bxd4  19. Qxd4  19... Rfe8  20. Rxe8+  20... Rxe8  21. Be3  21... b6} \\[-0.15cm]
        \tiny{22. Rc1  22...''}\\
        \tiny{\textbf{Label:}  ``  Qxc1+''}
        \end{tabular} 
        &
        \hspace{-0.5cm}
        \begin{tabular}{l}
                  
        \tiny{\textbf{Prompt:}  ``1. e4  1... e5  2. Nc3  2... Nf6  3. Nf3  3... Nc6  4. Bb5  4... Bc5  } \\[-0.15cm]
        \tiny{5. Bxc6  5... dxc6  6. d3  6... Bg4  7. h3  7... Bxf3  8. Qxf3  8... O-O  9. g4  }\\[-0.15cm]
        \tiny{9... Bb4  10. Bd2  10... Nd7  11. h4  11... Be7  12. g5  12... Nc5  13. O-O-O} \\[-0.15cm]
        \tiny{13... Qd7  14. h5  14... Qd8  15. Qg3  15... Ne6  16. Rdg1  16... b5  17. Qxe5} \\[-0.15cm]
        \tiny{17... a5  18. f4  18... Re8  19. Qf5  19... b4  20. Na4  20... Nd4  21. Qg4  21... c5} \\[-0.15cm]
        \tiny{22. f5  22... Ra6  23. f6  23... Bd6  24. fxg7  24... Kxg7  25. Rg2  25... Qc8  } \\[-0.15cm]
        \tiny{26. h6+  26... Kg8  27. Qh5  27... Qd7  28. Rf1  28... Re6  29. Rgf2  29... Rg6  } \\[-0.15cm]
        \tiny{30. c3  30... bxc3  31. Nxc3  31... a4  32. Nd5  32... Qb5  33. Nf6+  33... Kh8} \\[-0.15cm]
        \tiny{34. Qh3  34... Rb6  35. Be3  35... Ne6  36. Nxh7  36... Qxd3  37. Rd1  37... Qc4+  } \\[-0.15cm]
        \tiny{38. Kb1  38... Qxe4+  39. Ka1  39... Be5  40. Nf6  40... Qc4  41. Nd5  41... Rb7  42.''}\\
        \tiny{\textbf{Label:}  ``  Qf5''}
        \end{tabular} 
        \\[0.3cm]
        \hspace{-0.5cm}(a) Elo-695 puzzle &
        \hspace{-0.5cm}(b) Elo-2253 puzzle
    \end{tabular}
    \caption{
    \textbf{Chess puzzles: example datapoints.} Two representative examples of an easy \textbf{(a)} and a hard \textbf{(b)} chess puzzle with corresponding prompts and target label formats.
    }
    \label{fig:chess-data}
\end{figure}

\textbf{Data preprocessing.}\quad
The GPT-4 pretraining dataset included chess games in the format of move sequence known as Portable Game Notation (PGN).
We note that only games with players of Elo 1800 or higher were included in pretraining.
These games still include the moves that were played in-game, rather than the best moves in the corresponding positions. 
On the other hand, the chess puzzles require the model to predict the best move.
We use the dataset originally introduced in \citet{schwarzschild2021can} which is sourced from \url{https://database.lichess.org/\#puzzles}~\citep[see also][]{schwarzschild2021datasets}.
We only evaluate the models ability to predict the first move of the puzzle (some of the puzzles require making multiple moves).
We follow the pretraining format, and convert each puzzle to a list of moves leading up to the puzzle position, as illustrated in \Cref{fig:chess-data}.
We use $50k$ puzzles sampled randomly from the dataset as the training set for the weak models and another $50k$ for weak-to-strong finetuning, and evaluate on $5k$ puzzles. 
For bootstrapping (\Cref{sec:bootstrapping}), we use a new set of $50k$ puzzles from the same distribution for each step of the process.

\textbf{Training hyperparameters.}\quad We train (finetune) all models for 5 epochs using a batch size of 32.
We do not apply early-stopping.

\textbf{Weak labels.}\quad We produce weak labels by sampling predictions at temperature $T=0$ (greedy decoding) from the weak model on a held-out set of additional $50k$ puzzles. The weak labels are completions showing the highest likelihood move according to the weak model.

\textbf{Evaluation.}\quad To evaluate the models, we sample completions at temperature $T=0$ on the held out test set, and compute the fraction of datapoints where the model outputs the correct next move.

\begin{figure}
    \centering
    \includegraphics[width=1\linewidth]{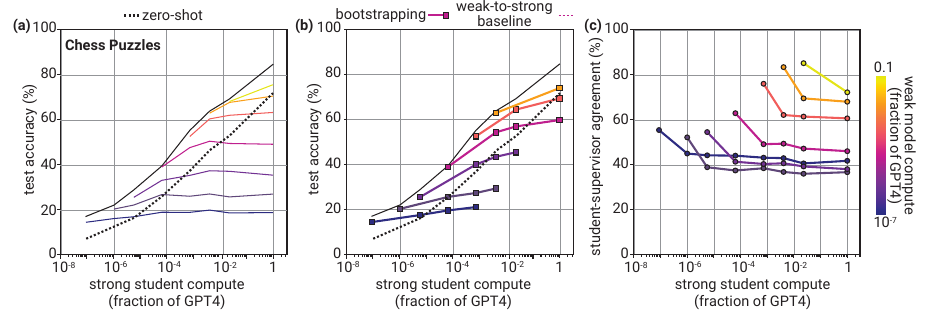}
 
\caption{
    \textbf{Additional results on chess.}
    Test accuracy of \textbf{(a)} baseline and \textbf{(b)} bootstrapping (see \cref{sec:bootstrapping}) compared to a zero-shot baseline. 
    Zero-shot performance improves with model size, and students supervised with much weaker supervisors sometimes underperform compared to the corresponding zero-shot model. 
    \textbf{(c)} Supervisor-student agreement on the chess puzzle data.
    Similar to \Cref{fig:agreement_breakdown}, agreement decreases as the student becomes larger.
    Hue of line indicates compute of weak supervisor.
    } 
\label{fig:chess_extra_results}
\end{figure}

\textbf{Zero-shot results.}\quad
In \Cref{fig:chess_extra_results}(a, b), we compare the naive baseline and bootstrapping (see \cref{sec:bootstrapping}) generalization to a zero-shot baseline on the chess puzzle data. 
Especially since the models were pretrained on chess games, zero-shot evaluation provides a  strong baseline.
In particular, strong students trained with much weaker supervisors underperform the zero-shot baseline for the same model size in some cases.

\textbf{Supervisor-student agreement results.}\quad
In \Cref{fig:chess_extra_results}(c), we report the supervisor-student agreement on the chess puzzles.
Similar to the NLP tasks (see \Cref{subsec:emulating-weak}), the agreement on chess also decreases as the student models get larger.

\subsection{ChatGPT Reward Modeling}

\textbf{Data preprocessing.}\quad
Each datapoint presents a dialog $d$ between a user and an assistant, with a last message coming from the user;
for each dialog, there are multiple candidate completions $(c_1, c_2, \ldots, c_m)$, i.e. responses from the assistant.
We also have access to pairwise comparisons of completions, where the labeler specifies the preferred completion within a given pair of completions.
To sum up, the datapoints can be viewed as $(d, c_1, c_2, y)$, where the label $y$ is $1$ if the labeler preferred completion $c_2$ and $0$ otherwise.  We use a mixture of multiple datasets used to train the reward models for ChatGPT. 

\textbf{Models.}\quad To adapt the language models to the reward modeling setting, we replace the unembedding layer of the model with a linear head with a single output, which is the logit for a given completion. The weights for this head are initialized to the unembedding weights of an arbitrary token in the original embedding layer. Similar to past work~\citep{stiennon2020learning,ouyang2022training}, we run two forward passes for each comparison, and the model prediction is given by $\sigma(\mathcal{M}_w(d, c_2) - \mathcal{M}_w(d, c_1))$,
where $\sigma$ is the sigmoid function and $\mathcal{M}_w(d, c)$ is the logit for completion $c$ predicted by the model.

\begin{figure}
    \centering
    \includegraphics[width=1\linewidth]{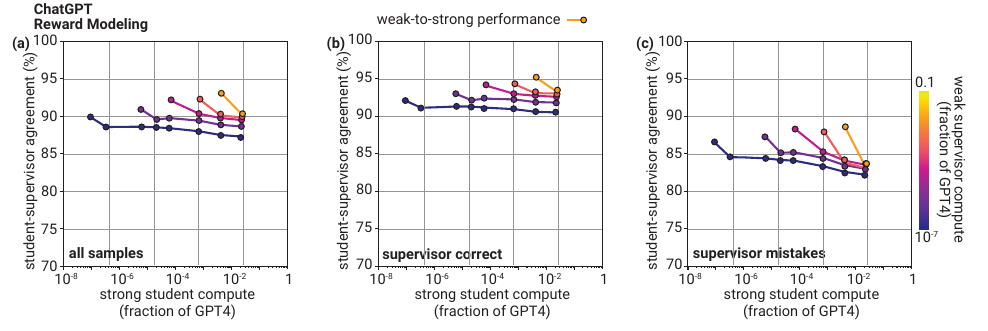}
 
\caption{
    \textbf{Supervisor-student agreement decreases for stronger students on RMs.}
    Please refer to caption of \Cref{fig:agreement_breakdown} for  detailed explanation of the plot.
    We reproduce the supervisor-student agreement experiment on the reward modeling data, and observe similar trends to the NLP tasks.
    } 
\label{fig:agreement_breakdown_rm}
\end{figure}

\textbf{Training hyperparameters.}\quad We train for 1 epoch with a batch size of $220$. We do not apply early-stopping.

\textbf{Weak labels.}\quad
We train the weak models on half of the available comparison data, and then make predictions on the other half. 
The weak label $y_w$ for a comparison $(d, c_1, c_2)$ is given by $y_{w} = \sigma(\mathcal{M}_w(d, c_2) - \mathcal{M}_w(d, c_1))$,
where $\sigma$ is the sigmoid function and $\mathcal{M}_w(d, c)$ is the logit for completion $c$ predicted by the weak model.

\textbf{Supervisor-student agreement results.}\quad In \Cref{fig:agreement_breakdown_rm}, we report the supervisor-student agreement on the RM task. 
Similar to the NLP tasks in \Cref{fig:agreement_breakdown} and chess puzzles in \Cref{fig:chess_extra_results}(c), the agreement decreases as the student gets larger.

\begin{figure}
    \centering
    \includegraphics[width=0.8\linewidth]{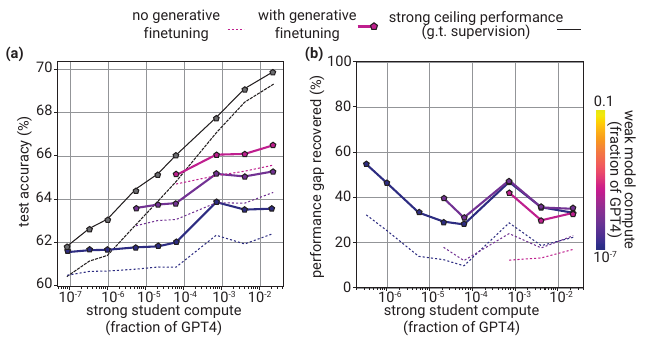}
    \caption{
    \textbf{The benefits of improved task-specific tuning and ground truth early stopping stack, resulting in even higher PGR.} Like \Cref{fig:rm_sft} but with ground truth early stopping based on test accuracy.
    }
    \label{fig:rm_sft_early_stop}
    \vspace{-10pt}
\end{figure}

\begin{figure}[t]
    \centering
    \includegraphics[width=0.8\linewidth]{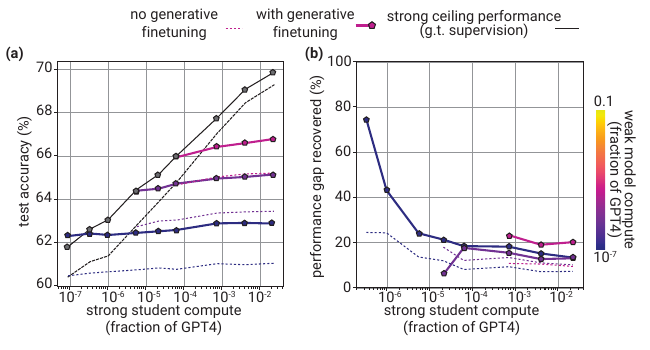}
    \caption{
    \textbf{PGR improves when \textit{both} supervisors and students have an extra generative fine-tuning step.}  Like \Cref{fig:rm_sft} but where ``with generative finetuning'' indicates that both supervisors and students have an extra generative finetuning step.  In other words, for this experiment all base models have an extra generative finetuning step following pretraining.
    }
    \label{fig:rm_sft_student}
    \vspace{-10pt}
\end{figure}

\textbf{Generative finetuning.}\quad
In \Cref{fig:rm_sft_early_stop}, we show that the PGR improvements from the generative finetuning on RM data (\Cref{sec:rm-sft}) and from early-stopping on ground truth test accuracy (\Cref{sec:results-overfitting}) stack together, leading to results competitive with the NLP and chess settings.
In \Cref{fig:rm_sft_student}, we report the results of an experiment similar to \Cref{fig:rm_sft}, but where the weak models are also pretrained with an additional generative finetuning step on the RM data.

\subsection{Auxiliary Confidence Loss}
\label{sec:app-aux-loss-details}

Here, we provide a detailed description of the method we use in \Cref{sec:auxloss}.

We use the following loss function:
\begin{equation}
    \label{eq:conf_loss}
    L_\text{conf} (f) = (1-\alpha) \cdot \text{CE}(f(x), f_{w}(x)) + \alpha \cdot \text{CE}(f(x), \hat{f}_t(x))
\end{equation}
where $\text{CE}(\cdot, \cdot)$ is the cross-entropy loss between the predictive distributions on a given input $x$, $f_w(x) \in [0, 1]$ represents the weak label predictive distribution, $f(x) \in [0, 1]$ is the strong model predictive distribution, $\alpha$ is a weight and $t$ is a threshold.
The predictions $\hat{f}_{t}(x)$ correspond to hardened strong model predictions using a threshold $t$, i.e.~$\hat{f}_{t}(x) = I[f(x) > t] \in \{0, 1\}$ where $I$ is the indicator function.
We set the threshold $t$ adaptively, so that $f(x) > t$ holds for exactly half of examples in the batch\footnote{The choice of exactly half reflects the prior over classes, and should be computed explicitly from weak model predictions in non-balanced or non-binary settings.}.
We set $\alpha_{\max} = 0.75$ for the largest student models and to $0.5$ otherwise and linearly warm-up $\alpha$ from 0 to $\alpha_{\max}$ over the first $20\%$ of training.

Our balancing mechanism incorporates a prior over the distribution of labels into training and is only practically feasible in the low-$n$ classification setting. For most weak-strong pairs and datasets, it had a small or neutral effect on weak-to-strong generalization; however, in a few settings it made a significant improvement.

We note that the loss in \Eqref{eq:conf_loss} can be rewritten as a self-bootstrapping loss:
\begin{equation}
    \label{eq:conf_loss_bst}
    L_\text{conf} (f) = \text{CE}(f(x), (1-\alpha) \cdot f_{w}(x)  + \alpha \cdot \hat{f}_t(x)),
\end{equation}
i.e. the cross-entropy target is a mixture of the weak model predictions and the (thresholded) predictions of the strong student itself.
This loss is related to the bootstrapping methods in~\citet{reed2014training} and \citet{arazo2019unsupervised} for addressing label noise.
It is also similar to self-training~\citep{lee2013pseudo} and conditional entropy minimization~\citep{grandvalet2004semi}, which have led to state-of-the-art results in semi-supervised learning~\citep{xie2020self} and domain adaptation~\citep{shu2018dirt}.
\citet{chen2020self} and  \citet{wei2020theoretical} show that self-training can mitigate the bias of the supervisor model.

In \Cref{sec:app-other-methods} we also describe other methods we considered; for most of these methods, we got negative early results.

\section{Additional results on methods}
\label{sec:app-other-methods}

\begin{table}[ht]
    \caption{
    \textbf{Custom prompts used in the zero-shot and few-shot experiments}.
    We design a simple custom prompt for each of the tasks in the table below.
    In the few-shot setting, we also append labeled (with ground truth or weak labels) examples to the prompt.
    }   
    \label{tab:prompts}
    \centering
    \begin{tabular}{cc}
    \toprule
    Dataset & Custom Prompt \\
    \midrule \midrule
        CosmosQA & 
        \begin{tabular}{l}
             \it{\small There is some context, followed by a question about the context, and an answer.} \\
             \it{\small Respond with 1 if the answer is correct, and 0 otherwise.}
        \end{tabular}
    \\[0.3cm]
        Deontology & 
        \begin{tabular}{l}
             \it{\small There is a short scenario, consisting of a situation or request, followed by an excuse. } \\
             \it{\small Respond with 1 if the excuse is valid, and with 0 otherwise.} 
        \end{tabular}
    \\[0.3cm]
        Justice & 
        \begin{tabular}{l}
             \it{\small There is a statement. Respond with 1 if it is correct or just and with 0 otherwise. }
        \end{tabular}
    \\[0.3cm]
        QuAIL & 
        \begin{tabular}{l}
             \it{\small There is a news context, followed by a question about the context, and an answer} \\
             \it{\small to the question. Respond with 1 if the answer is correct, and with 0 otherwise.} 
        \end{tabular}
    \\[0.3cm]
        SciQ & 
        \begin{tabular}{l}
             \it{\small There is a science knowledge question, followed by an answer. } \\
             \it{\small Respond with 1 if the answer is correct, and with 0 otherwise.} 
        \end{tabular}
    \\[0.3cm]
        Social IQa & 
        \begin{tabular}{l}
             \it{\small There is some context, followed by a social question, followed by an answer. } \\
             \it{\small Respond with 1 if the answer is correct, and 0 otherwise.} 
        \end{tabular}
    \\[0.3cm]
    Virtue & 
        \begin{tabular}{l}
             \it{\small There is a short scenario, followed by a judgement of the person involved. } \\
             \it{\small Respond with 1 if the judgement is correct, otherwise respond with 0.} 
        \end{tabular}
    \\
    \bottomrule
    \end{tabular}
\end{table}

We did preliminary experiments on a variety of methods for improving the strong model performance in our weak-to-strong generalization setting. 
We found many of them not useful for improving over the naive finetuning baseline, and others yielding limited improvements on a subset of settings but not consistently over all datasets and model sizes. 
We summarize the algorithms, the motivations, and the takeaways below. 
Note that we did not optimally tune each of the methods, so it is possible that with better tuning they may still perform well.

\textbf{Confidence thresholding.}\quad To filter out incorrect weak labels, we used a simple cut-off method that selected only the top $5\%$ to $20\%$ examples from each class where the weak supervisor is most confident to train the strong model. 
We found that our weak labels are typically well-calibrated, but confidence thresholding only helps when the weak labels are very bad (e.g.~60\% accuracy) and stops being useful when the weak labels reach around 70\% to 80\% accuracy.
We observed these results both in NLP and in the chess puzzle settings. 
See \Cref{sec:app-easy-hard} for more discussion of related experiments.

\textbf{Confidence losses.}\quad To encourage strong model to make confident predictions ~\citep{grandvalet2004semi}, we added an auxiliary loss that encourages the model predicted class probability $p$ to be far away from 0.5. We tried both the $l_2$ loss $-(p-0.5)^2$ and the entropy loss $p \log p + (1-p) \log (1-p)$. 
We found these losses to be helpful in preliminary experiments in the linear probing setting, but they generally performed less well than the confidence auxiliary loss in Equation~\ref{eq:conf_loss} in the finetuning setting.
We have also observed negative results with the confidence losses when the training data is highly class-imbalanced or when we do not use the rebalancing procedure described in \Cref{sec:method}.

\textbf{Product confidence loss.}\quad We also tried a confidence-like loss which sets the cross entropy targets to be proportional to the product of the probabilities that the weak and strong models assign, renormalized across classes and without propagating gradients through the targets.  In preliminary experiments, this loss consistently gave positive results over the baseline on two NLP tasks, but performed poorly compared to our main confidence loss.  Variants like geometric mean instead of product gave no boost.  Compared to the confidence loss, it could be useful as it has no inter-batch dependence and could potentially be adapted for generative tasks.

\textbf{LP-FT.}\quad We used the LP-FT technique proposed in~\citet{kumar2022fine} which first trains a linear probe on frozen strong model representations and then finetunes all layers, to avoid destroying the pretrained representation. 
We were unable to get improvements compared to the finetuning baseline.

\textbf{Weight regularization.}\quad To regularize the strong model weights to avoid imitating the weak labels\footnote{However, as we discuss in \Cref{subsec:emulating-weak}, in our setup the strong model tends to be bad at imitating the weak labels. Therefore, regularization could be more important in settings where the strong model can fit the weak labels well.}, we tried a variety of regularization techniques for strong model training, including stronger weight decay~\citep{krogh1991simple} and dropout~\citep{srivastava2014dropout}. 
We did not find significant improvement. 

\textbf{LoRA.}\quad As another regularization technique, we also considered low-rank adaptation (LoRA) ~\citep{hu2022lora}, i.e. only making a low-rank update to the parameters of each layer of the model during finetuning. We did not find any improvement, even when sweeping the LoRA rank.

\textbf{Data augmentation.}\quad Inspired by the success of consistency algorithms in self-supervised training~\citep{chen2020big, caron2021emerging}, we used the strong student models to rephrase the inputs in each sample, and added an auxiliary loss enforcing the strong model predictions to be consistent between original and rephrased samples. We did not find any improvement on a selected subset of NLP datasets.

\textbf{Adding label noise, special losses for noisy labels.}\quad 
We experimented with the generalized cross-entropy loss proposed in~\citet{zhang2018generalized} that is more robust to label noise, but did not find improvement over cross-entropy. We also tried adding random noise to weak labels, and found that the strong models were able to simulate the weak labels less well, especially early in training, but it did not ultimately result in improved performance.

\textbf{Few-shot prompting.}\quad As an alternative to fine-tuning, we can use the in-context learning ability of the strong student models.
For each task, we append a custom prompt shown in \Cref{tab:prompts}.
For a detailed description of the results, see \Cref{subsec:prompting}.

\textbf{Weight averaging.}\quad
Prior work~\citep{izmailov2018averaging, cha2021swad, wortsman2022robust, wortsman2022model} suggested that various forms of weight averaging can substantially improve performance, especially in distribution shift settings.
In our setup, we experimented with applying exponential moving averaging to the parameters of the model during training, but did not observe improvements relative to the baseline.

\section{Easy-to-hard generalization}
\label{sec:app-easy-hard}

In \Cref{subsec:emulating-weak} and \Cref{sec:app-simulation}, we discuss that one reason weak-to-strong generalization may be difficult is if the weak labels have systematic errors that the strong model can learn to emulate.  One natural type of systematic weak label error is to do poorly on hard examples and well on easy examples.

In this section, we focus on studying what we call \textbf{easy-to-hard generalization}, where we train only on easy examples using ground truth supervision, and assess generalization to harder examples.

\subsection{Chess puzzles}
\label{sec:app-chess}

\begin{figure}
    \centering
    \includegraphics[]{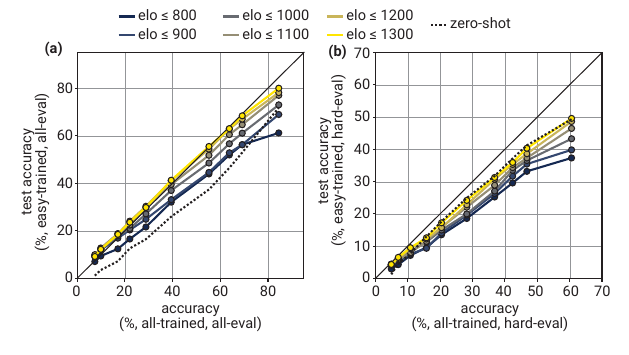}
    \caption{
    \textbf{Easy-to-hard generalization on chess puzzles.}
    We finetune models on chess puzzles with Elo $\le t$, varying the threshold $t$, and evaluate the finetuned models on 
    \textbf{(a)}: all test puzzles, and \textbf{(b)}: hard test puzzles with Elo $\ge 2000$.
    Across the board, we see strong performance, even when training only on very easy puzzles (Elo $\le 800$).
    For reference, we also include the zero-shot performance of the model.
    Finetuning on easy puzzles, we improve upon the performance on average on the test set, but we do not improve on hard puzzles, compared to the zero-shot model.
    }
    \label{fig:chess-eh}
\end{figure}

\begin{figure}
    \centering
    \begin{tabular}{c}
        \includegraphics[width=0.95\linewidth]{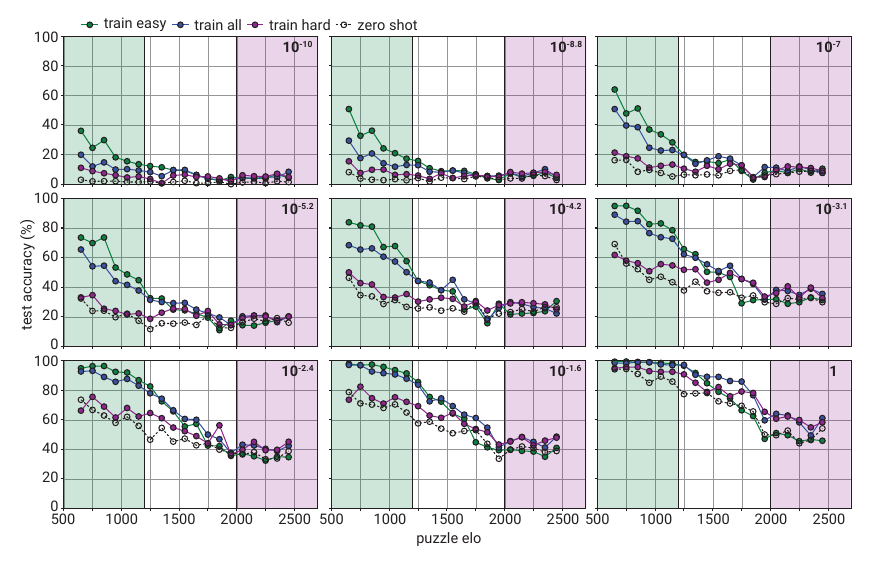} \\
        (a) Easy cutoff: Elo $\le$ 1200 \\[0.3cm]
        \includegraphics[width=0.95\linewidth]{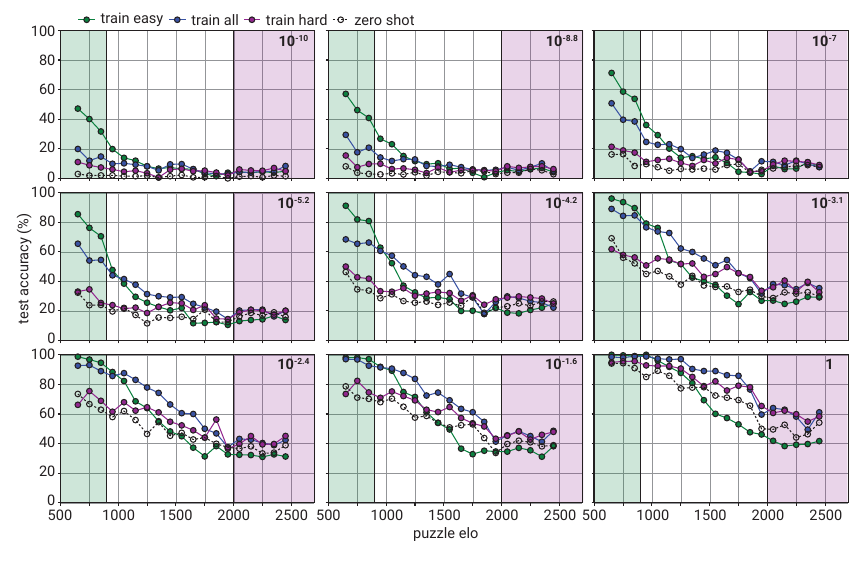} \\
        (b) Easy cutoff: Elo $\le$ 900
    \end{tabular}
    \caption{
    \textbf{Easy-to-hard generalization on chess puzzles.}
    We present detailed performance of models finetuned on different subsets of chess puzzles across model sizes and test puzzle difficulty levels.
    For each model size, we compare models trained only on easy puzzles, hard puzzles, or all puzzles.
    We also include the zero-shot model performance.
    We provide results for the easy puzzle Elo cutoffs of \textbf{(a)}: 1200 and \textbf{(b)}: 900.
    All finetuned models are trained on $50k$ random datapoints from the corresponding distribution. 
    The size of the model is shown in the upper-right corner of each panel, in terms of fraction of GPT-4 compute.
    }
    \label{fig:chess-eh-detailed}
\end{figure}

Each chess puzzle comes with a natural difficulty label:  an Elo score, which describes its difficulty according to humans.  On the \url{https://lichess.org} website, people try to solve puzzles, which can be viewed as a game between a puzzle and a human player. 
The Elo scores are then assigned to both human players and chess puzzles following the standard Elo algorithm.

We consider the easy-to-hard generalization problem, where the difficulty is defined according to the puzzle Elo rating.
We note that the puzzle Elo describes the difficulty of the entire puzzle move sequence, while we are only training the model to predict the first move in the sequence (see \Cref{sec:app-details-chess}).
Consequently, the puzzle Elo is a high-quality but still imperfect measure of difficulty of the problem for humans.
It is also important to note, that puzzle Elo may not be a good measure of difficulty for the models: easy puzzles for humans can be hard for the models and vice versa.

We then split the dataset into subsets according to the puzzle Elo.
We consider the hard set to be puzzles with difficulty above Elo 2000.
For the easy set, we consider cuttoffs in $\{800, 900, 1000, 1100, 1200, 1300\}$, and use puzzles with difficulty below the cutoff.
We also consider the unrestricted set of \textit{all} puzzles. 
We sample $50k$ puzzles from each of these sets randomly, and finetune the model on them\footnote{For easy puzzles with 800-Elo cutoff, we only use $25k$ puzzles, because there are not $50k$ puzzles available in this difficulty range.}.

We report the results in Figure \ref{fig:chess-eh}, where we also provide the performance of a zero-shot baseline for reference.
We plot the accuracy of the models trained on the easy subsets of puzzles against the performance of the same model trained on all puzzles.
We find that the models generally perform well on average on the test set in panel (a), and outperform the zero-shot baseline.
Interestingly, when evaluated on hard examples only, in panel (b), the models perform similarly to the zero-shot baseline, or slightly worse.

When trained on easy puzzles, the models shift towards performing well on the easy puzzles, and underperform on the hard puzzles. 
In Figure \ref{fig:chess-eh-detailed}, we can see that generally the models improve upon the zero-shot baseline outside of their training difficulty range, often up to Elo of 1500 or higher, but underperform on the hardest examples.

\begin{figure}
    \centering
    \includegraphics[width=0.95\linewidth]{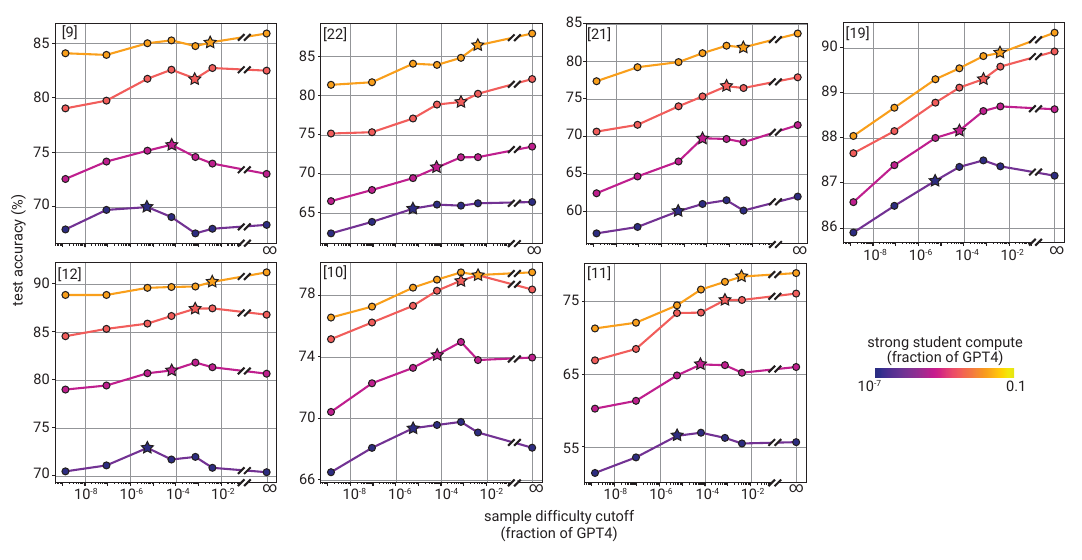}
    \caption{\textbf{Effect of varying training data difficulty on test set accuracy.}
    Test accuracy as a function of sample difficulty cutoff on a subset of our NLP tasks. 
    The leftmost point on the horizontal axis corresponds to only using datapoints that models of all sizes that we consider get right when trained on other data sampled from the same task, and the rightmost point (denoted with $\infty$) corresponds to training on all datapoints; the point with value $x$ on the horizontal axis corresponds to only using the datapoints that models with $x$ or higher compute (fraction of GPT-4) consistently get right. 
    Inset numbers indicate task id (compare \Cref{fig:all_grid_nlp}). Hue indicates compute of weak supervision. Stars indicate points where weak supervisor size corresponds to sample difficulty cutoff.
    }
    \label{fig:difficulty-thresholding}
\end{figure}

\subsection{NLP tasks: difficulty thresholding}

NLP tasks do not come with a natural source of difficulty labels, but we can create such labels by looking at performance as a function of model size.

We define \textit{difficulty} of a datapoint based on the smallest model size that consistently predicts the label on this datapoint correctly, when trained on ground truth. 
For example, suppose we have 4 ground truth models $W_1$, $W_2$, $W_3$, $W_4$ that use compute $C_1 < C_2 < C_3 < C_4$ respectively. 
Suppose models $W_1$, $W_3$, $W_4$ predict the example correctly when it is in a held-out set, while $W_2$ predicts it incorrectly. Then we will assign a difficulty of $C_3$ to the example.

Then given a difficulty cutoff $D$, we filter the training set to examples with difficulty $\leq D$.
We subsample the filtered set so that the number of training examples is equal to the number of examples at the lowest difficulty level.
We train a model on the subsampled training set using ground truth labels, and measure its accuracy on a held out test set (with no subsampling).

The subsampling ensures that we use the same training set size for each difficulty cutoff. 
Using ground truth labels ensures that the label accuracy is the same ($100\%$) for each cutoff. We also use the same test set for each cutoff. This setup lets us vary only training data difficulty, and measure its impact on the trained model's accuracy.

We plot results in \Cref{fig:difficulty-thresholding}. The $y$-axis is accuracy on the test set, while the $x$-axis is the difficulty cutoff. 
Increasing the difficulty cutoff generally leads to an increase in accuracy. 
This result suggests that solving easy-to-hard generalization is non-trivial even if there are no weak label errors.

For smaller models (darker lines), the accuracy initially increases, but starts to decrease beyond a point. 
The drop generally happens when the difficulty cutoff exceeds the capacity of the model itself, i.e. when the examples are too difficult for the model to fit. 
However, large models trained on easy examples often perform well.

\begin{figure}
    \centering
    \includegraphics[]{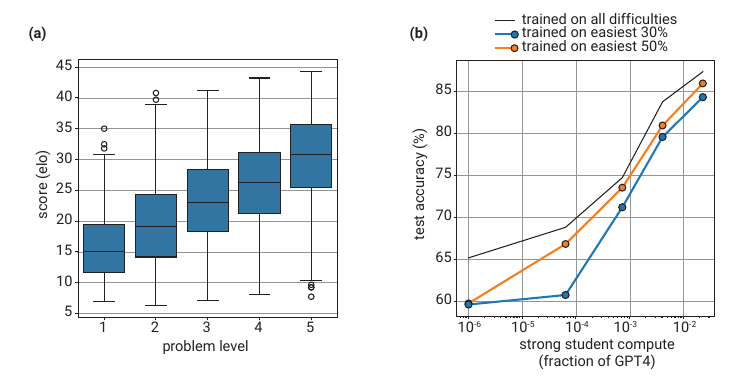}
    \caption{\textbf{Filtering training samples by GPT-4 generated Elo scores results in very good easy-to-hard generalization.} \textbf{(a)} GPT-4 generated Elo scores for different, human-defined, problem difficulties (1 - easiest, 5 - hardest) on the MATH dataset. \textbf{(b)} Average test accuracy as a function of strong student compute on a subset of our NLP tasks. Student is trained on ground truth labels on samples of all difficulties (black), only the 30\% easiest tasks (orange), or only the 50\% easiest tasks (blue). }
    \label{fig:elo_math}
\end{figure}

\subsection{GPT-4 predicted difficulty}

Ultimately, we care about strong models generalizing from human supervision.
From this perspective, it is important to understand whether we can achieve easy-to-hard generalization, where the difficulty is measured according to humans, rather than capacity-constrained models.
In \Cref{sec:app-chess}, we explored this question in chess, but we would want to extend this analysis to the NLP tasks.

Most natural datasets do not come with information about problem difficulty.
As a rough estimate, we automatically generated difficulty labels using GPT-4.
More concretely, we used GPT-4 to rank pairs of examples in each dataset, asking ``which question is easier, Question A or Question B?'' We then calculated the Elo scores for each example via a finite number of random comparisons.

To evaluate the quality of GPT-4 Elo score as a measure of difficulty, we performed correlation analysis against human annotations for datasets with human difficulty levels such as MATH~\citep{hendrycks2021measuring} and chess, as well as against weak model confidence. 
We found that the three measures align better for reasoning tasks such as MATH, as we show in \Cref{fig:elo_math}(a), but not much for some natural language tasks. 
When looking at the samples, we found that GPT-4 Elo scores tend to be higher for longer questions, but those questions may actually be easy for smaller models since they provide more context.

Using GPT-4 Elo score as a proxy for human difficulty, we used different cutoffs on scores to separate easy and hard examples, trained the strong models on the easy examples only (with ground truth labels), and evaluated on the hard examples. Preliminary results are shown in \Cref{fig:elo_math}(b).

In general, we found that using GPT-4 Elo as measure of hardness makes generalization slopes steeper than our main setup of weak-to-strong generalization. 
One possible confounder for interpretation is that our Elo measurements could be noisy, causing generalization to be better. 

Note that this setup is a classic covariate shift problem, whereas our main setup focuses more on concept shift and noisy labels. 
It is unclear which setup would be more relevant, and we think it is important to study easy-to-hard generalization more thoroughly in future work.

\section{Other weak-to-strong settings}
\label{sec:app:more-weak-to-strong}

\subsection{Self-supervised vision models}
\label{app:sec:imagenet}

We additionally demonstrate weak-to-strong generalization in a simple image classification experiment.
We use a pretrained AlexNet model~\citep{krizhevsky2012imagenet} as a weak supervisor, and use it to generate weak labels on the ImageNet~\citep{russakovsky2015imagenet} validation set.
As a strong student, we use linear probing on frozen representations extracted by DINO models~\citep{caron2021emerging} based on ResNet-50~\citep{he2016deep} and ViT-B/8~\citep{dosovitskiy2020image} architectures.
The DINO models are pretrained in an unsupervised way and did not observe direct supervision for ImageNet classification or any other classification task during pretraining, so this experiment does not have the pretraining leakage disanalogy discussed in \Cref{subsec:disanalogies}.

\begin{table}[t]
    \caption{
    \textbf{Weak-to-strong generalization on ImageNet}.
    We train linear probes on the representations extracted by DINO models with weak supervision from an AlexNet model.
    The strong students substantially outperform their weak supervisor.
    }   
    \label{tab:imagenet}
    \centering
    \begin{tabular}{ccc}
    \toprule
    Model & Top-1 Accuracy (\%) & PGR (\%) \\
    \midrule \midrule
    AlexNet (weak supervisor)                 & 56.6           & - \\
    \midrule
    Dino ResNet50          & 63.7           & - \\
    Dino ViT-B/8           & 74.9           & - \\
    \midrule
    AlexNet $\rightarrow$ DINO ResNet50 & 60.7 & 57.8\\
    AlexNet $\rightarrow$ DINO ViT-B/8  & 64.2 & 41.5 \\
    \bottomrule
    \end{tabular}
\end{table}

We use $40k$ datapoints from the validation set to train the linear probes, and evaluate performance on the remaining $10k$ datapoints.
For training the linear probes, we use a batch size of $128$, Adam optimizer~\citep{kingma2014adam} and a learning rate of $10^{-3}$.
We run $20$ epochs of training for ResNet-50 and $5$ epochs for ViT-B/8.

We report the results in \Cref{tab:imagenet}.
Similarly to our main experiments in \Cref{sec:results}, the student can substantially outperform the supervisor, achieving PGR on the order of 50\%. 
This experiment shows that our results are not limited to the natural language setting, and generalize to other domains. 
It also shows that strong students can generalize from weak supervision on tasks where they only had indirect pretraining, i.e. where the knowledge of the task is latent. 

\subsection{Linear probing}
\label{sec:app-lp}

\begin{figure}[ht]
    \centering
     \includegraphics[width=0.9\linewidth]{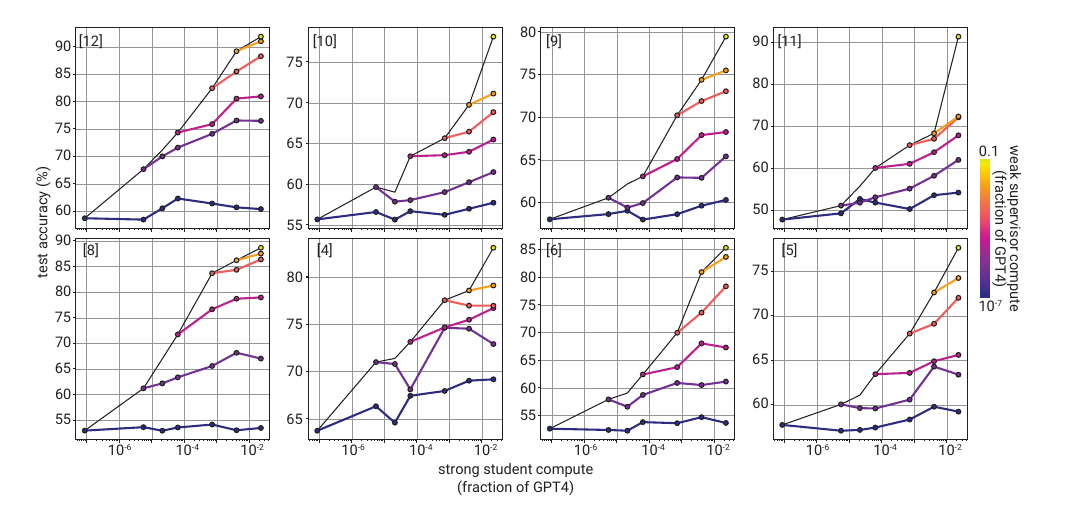}
    \caption{\textbf{Linear probing qualitatively matches finetuning weak-to-strong generalization.} Test accuracy as a function of strong student compute on a subset of our NLP tasks. Inset numbers indicate dataset id (compare \Cref{fig:all_grid_nlp}). Accuracy of a linear probe on student model trained with ground truth in black, accuracy of linear probe on students trained directly with weak linear probe supervision shown in solid lines with circles (hue indicates compute of weak supervision).}
    \label{fig:lp}
\end{figure}

In addition to our main finetuning experiments, we also perform weak-to-strong generalization experiments in the linear probing setting.
We freeze all weak and strong model parameters, and train new linear classification heads both using ground truth labels and using weak labels. 
We train linear probes with Adam optimizer~\citep{kingma2014adam}, $10^{-3}$ learning rate, batch size 128, and no weight decay for 200 epochs, for both weak and strong model training.
We do early stopping based on agreement to the weak labels on the validation set and report test accuracy.
Results are shown in Figure~\ref{fig:lp}. 
We observe qualitatively similar generalization compared to the full finetuning case.

Generally, we found the linear probing setting to be very useful to quickly iterate on methods, datasets and ideas. 
While finetuning provides better results, the qualitative trends in linear probing are similar, and the experiments are much faster and easier to run.
For example, we initially found positive results with confidence loss (\Cref{sec:method}) and bootstrapping (\Cref{sec:bootstrapping}) in the linear probing setting.
\section{The effects of weak label structure}
\label{sec:app-simulation}

One challenge in weak-to-strong generalization is the presence of errors in the weak labels.
Throughout most of this paper, we consider a particular type of weak error structure: the kinds of errors smaller, capacity-constrained language models make. 
However, this is not the only type of errors possible.

In this section, we analyze synthetic examples of other kinds of weak label structures, and the implications they have on generalization.
Weak model error structure must be considered in relation to the particular strong model at hand. For example, we conjecture that the extent to which the strong model can imitate the weak supervisor may be very important. If we have two strong models of the same performance on the actual task but one is very good at imitating the labels, then we expect that model will generalize less desirably, at least with the naive finetuning method.

In \Cref{subsec:emulating-weak} we found that surprisingly the strongest students are imitating the weak supervisor mistakes less than smaller student models in our setting. Since we expect superhuman models to be very good at imitating human supervisor, this may be a major disanalogy. 
In this section we test cases where the weak supervisor can be imitated easily.

\subsection{Synthetic experiments on simulation  difficulty}

\begin{figure}
    \centering
    \includegraphics[width=0.9\linewidth]{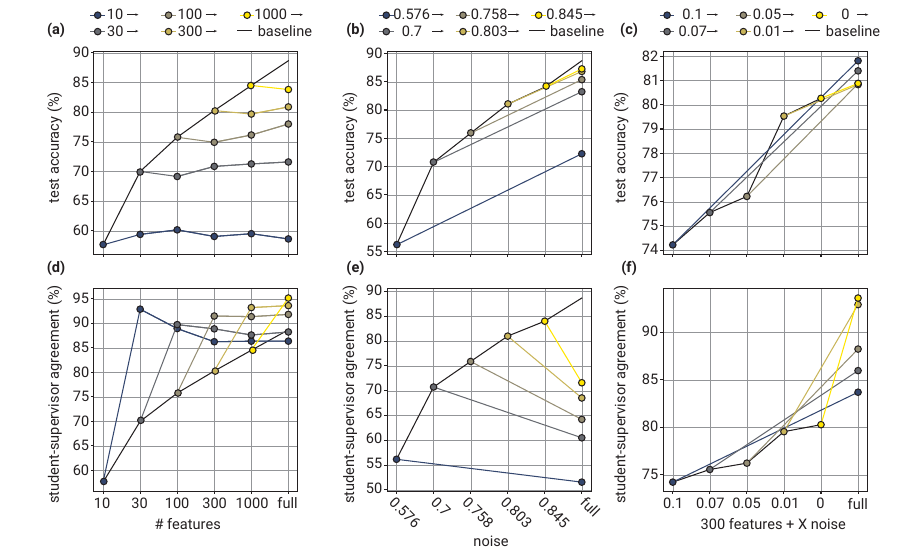}
    \centering
    \caption{
    \textbf{Synthetic experiment on simulation difficulty.}
    We consider three types of weak errors in a linear probing setting: \textbf{(a,d)} perfectly simulatable, where weak models use a subset of strong model features; \textbf{(b,e)} completely unsimulatable, where the weak labels are obtained by applying random noise to the ground truth; \textbf{(c,f)} a mixture of the two settings, where label noise is applied to perfectly simulatable weak labels.
    Top row of panels shows test accuracy and bottom row shows agreement to the weak labels.
    In addition to weak label accuracy, the structure of mistakes plays a major role in weak-to-strong generalization.
    }
    \label{fig:app-simulator}
\end{figure}

First, we consider a simplified linear probing setting, where we can ensure that the student can perfectly simulate the supervisor predictions by construction.
Specifically, we extract a representation $X \in \mathbb{R}^{n \times d}$ of the SciQ dataset using a model of an intermediate size in the GPT-4 family, where $n$ is the number of datapints, and $d$ is the dimensionality of the residual stream~\citep{elhage2021mathematical}.
We can then consider the family of linear models\footnote{We train logistic regression using the default parameters in the \lstinline{sklearn.linear_model.LogisticRegression} class~\citep{JMLR:v12:pedregosa11a} for this experiment.} $\mathcal{M}_k$ where $k \le d$ by training a linear probe only on the first $k$ features extracted by the model.
In particular, for $k=d$ we recover the standard linear probe.
By construction for $k_1 \ge k_2$, the model $\mathcal{M}_{k_1}$ can perfectly simulate $\mathcal{M}_{k_2}$.

Next, we can run our standard weak-to-strong generalization experiment, following the setup described in \Cref{sec:methodology}, using the family of models $\mathcal{M}_k$.
We train the weak supervisor models on $10k$ datapoints, and produce hard weak labels on the remaining $13k$ datapoints.
We report the results in \Cref{fig:app-simulator}(a,d). 
In this setting, the simulation is very easy, and we do not observe substantial improvements in the strong student model compared to the supervisor performance.
The test agreement values are substantially higher than the weak model accuracy, indicating that the students are overfitting to the supervisor errors. 
Interestingly, even in this simple setting the agreements are not $100\%$, likely due to the fact that the student models are trained on finite data, and with light $l_2$-regularization.

We can also consider the opposite setting: what if the student model cannot simulate the mistakes of the weak teacher at all?
Specifically, we generate weak labels by randomly flipping the labels to match the accuracy of the weak models from the previous experiment.
As a result, we get weak labels with the same accuracy, but which are completely unpredictable. 
In \Cref{fig:app-simulator}(b,e), when we train the student model on the these weak labels, we can get substantially higher accuracy than the accuracy of the weak labels.
In other words, if the errors of the weak supervisor are completely unpredictable (random) for the student, with enough data we should be able to recover good generalization, substantially exceeding the performance of the supervisor.

\begin{figure}[ht]
    \centering
    \begin{minipage}{0.6\textwidth}
        \centering
        \includegraphics[width=\linewidth,clip=true, trim={0 2.4cm 0 0}]{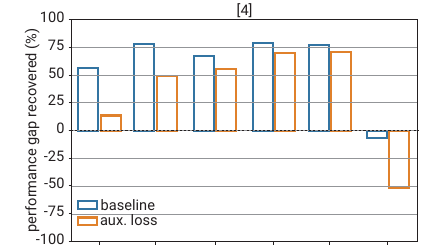}
        \label{fig:dream}
    \end{minipage}%
    \hfill
    \begin{minipage}{0.6\textwidth}
        \centering
        \includegraphics[width=\linewidth,clip=true, trim={0 2.4cm 0 0}]{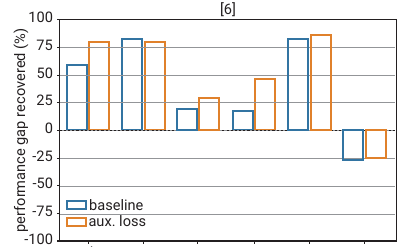}
        \label{fig:mctaco}
    \end{minipage}
    \hfill
    \begin{minipage}{0.6\textwidth}
        \centering
        \includegraphics[width=\linewidth]{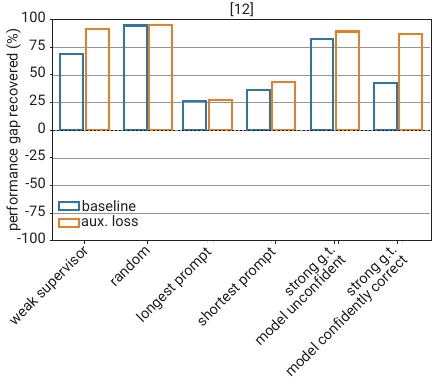}
        \label{fig:sciq}
    \end{minipage}
    \caption{\textbf{PGR for weak labels with same accuracy but different error structures.}     The inset number in each panel indicates the dataset (compare \Cref{fig:all_grid_nlp}).  Weak-to-strong generalization and methods both depend critically on the structure of the weak supervisor errors. While it is trivial to pick error structures that generalize well (for instance, random noise), these error structures are also very disanalogous to the ultimate superalignment setting, where we want to study the structures of human errors.
    }
    \label{fig:weak_labelers}
\end{figure}

\FloatBarrier

Finally, in \Cref{fig:app-simulator}(c,f) we consider a mixture of these two settings: 
we start with a perfectly simulatable weak model $\mathcal{M}_{300}$, and then add various amounts of label noise to the resulting weak labels. 
By training a strong student model (using all features) on the resulting weak labels, we recover the performance close to the performance of $\mathcal{M}_{300}$.

\paragraph{Discussion of results.}
The simple experiment in this section suggests that in addition to the weak label accuracy, it is important to consider the \emph{structure of weak errors}.
In particular, if the weak errors are extremely easy for the strong model to simulate, the student may not generalize much better than the weak supervisor with naive finetuning on the weak labels.
On the other hand, if the mistakes of the weak supervisor are completely unpredictable, the student can denoise the predictions of the supervisor and generalize better.
In future work, we believe it is important to consider various types of weak supervision with different structures of mistakes, and build a better understanding of how they affect weak-to-strong generalization.

\subsection{Different weak error structure means different generalization}

To further explore the impact of different weak error structures, we created several synthetic sets of weak labels for each dataset, all with error rate identical to the weak model's error rate.  To construct these labels, we start from ground truth, and then flip a subset of labels to match the accuracy of a particular weak model.  We target a few types of error structures, such as pure noise, easy-to-model bias, hard-to-model bias, and adversarial bias.

In particular, we looked at:

\begin{enumerate}
    \item \lstinline{weak supervisor}: the baseline --- labels are generated in the same way as in the rest of the paper
    \item \lstinline{random}: flip the label of random datapoints
    \item \lstinline{longest prompt}: flip the label of longest datapoints by characters
    \item \lstinline{shortest prompt}: flip the label of shortest datapoints by characters
    \item \lstinline{strong g.t. model unconfident}: flip the label of the datapoints that the strong ceiling model is most unconfident on
    \item \lstinline{strong g.t. model confidently correct}: flips the label of the datapoints that the strong ceiling model is most confidently correct on
\end{enumerate}

Despite all of these weak labelers having the same weak accuracy, we find that the generalization can vary wildly depending on the structure of the weak errors.
We report the results in \Cref{fig:weak_labelers}.

\begin{figure}
    \centering
        \begin{minipage}{0.95\textwidth}
        \centering
        \includegraphics[width=\linewidth]{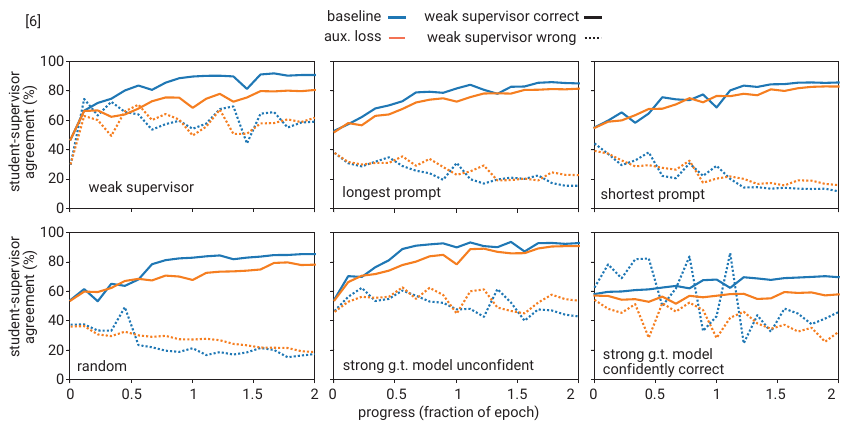}
    \end{minipage}%
    \hfill
    \begin{minipage}{0.95\textwidth}
        \centering
        \includegraphics[width=\linewidth]{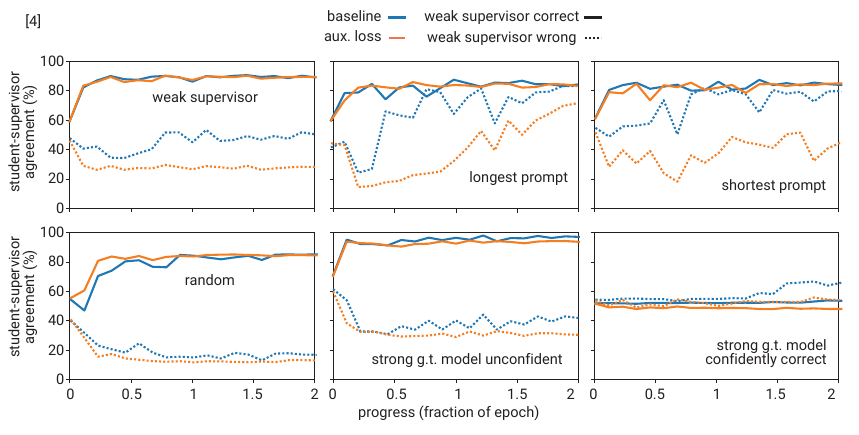}
    \end{minipage}
    \hfill
    \begin{minipage}{0.95\textwidth}
        \centering
        \includegraphics[width=\linewidth]{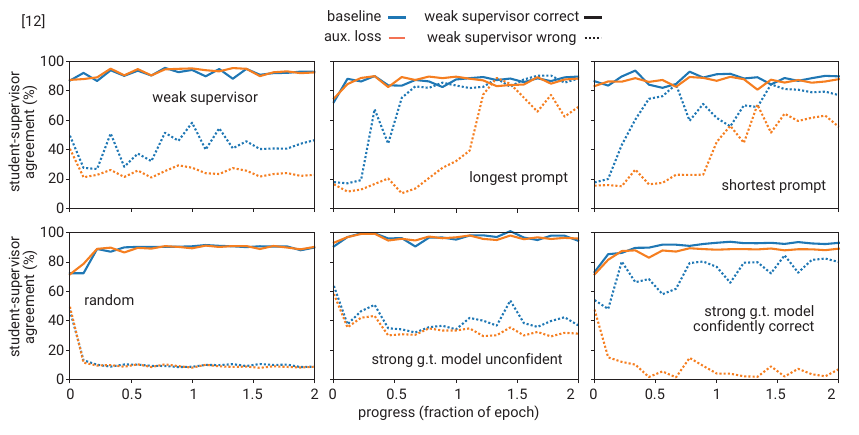}
    \end{minipage}
    
    \caption{\textbf{Training dynamics change for different weak errors.} We show teacher-student agreement for different weak error structures on three datasets. We see that the training dynamics have qualitatively different behavior for different error structures, despite all weak labelers having the same accuracy.}
    \label{fig:synthetic-imitation-dynamics}
\end{figure}

Furthermore, the dynamics of supervisor-student agreement through training can have qualitatively different behavior (\Cref{fig:synthetic-imitation-dynamics}).  
For errors coming from a weak model, we see that there is often initially a period of generalization, followed by a period of overfitting where it learns the weak model’s errors.  The confidence auxiliary loss mitigates this overfitting.
For easy-to-fit error structures such as \lstinline{longest prompt}, the overfitting happens much faster.  
For other kinds of errors, such as random noise, we often see that generalization improves throughout:  weak errors are not modeled, but the signal from the weak model is.

\begin{figure}
    \centering
    \includegraphics[width=0.75\linewidth]{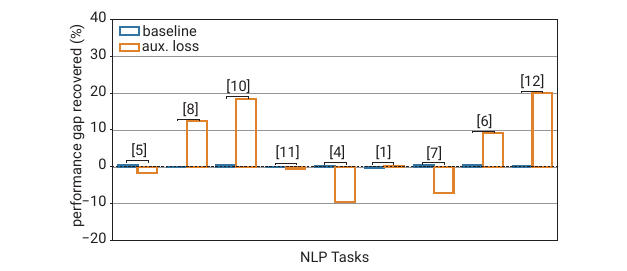}
    \caption{\textbf{Generalization when emulating weak labels is trivial.} Very little weak-to-strong generalization occurs if emulating the weak labels is trivial: average PGR across tasks is $0.002 \pm 0.003$ for baseline, and $0.046 \pm 0.108$ for aux loss, compared to around $0.2$ and $0.8$ respectively for the original tasks.}
    \label{fig:trivial-imitation}
\end{figure}

\subsection{Making imitation trivial}
\label{sec:app-imitation}

One possible major disanalogy in our setup, as discussed in \Cref{subsec:disanalogies}, is the fact that our models are not very good at imitating the weak model\footnote{Also known as learning the ``human simulator'' in the terminology of \citet{christiano2022eliciting}.} (\Cref{subsec:emulating-weak}), but superhuman models may be very good at imitating humans. It is possible that if the strong model were good at imitating the weak model, then it would generalize substantially less desirably by default.

To test an extreme version of this hypothesis, we create a synthetic setting where the strong model can trivially imitate the weak model very well. In particular, we modify the task by appending ``I think this is \texttt{\{weak\_label\}}. What do you think?'' to every prompt, where weak\_label is ``correct'' or ``incorrect'' based on the weak model prediction.
In this case, the hardened weak label is present in-context, and the simulation is trivial.

As expected, we find that both the baseline and the confidence loss introduced in \Cref{sec:method} show poor weak-to-strong generalization (\Cref{fig:trivial-imitation}) in most cases.
Interestingly, the confidence loss still improves upon the baseline achieving non-trivial generalization in several tasks.

\section{How should we empirically study superalignment, methodologically?}
\label{sec:app-how-study}

What makes a setup good for studying superalignment in the first place, all things considered? Tractability and ease of study are clearly important criteria, but also certainly not the only ones.
This question is non-obvious because superalignment is qualitatively different from other machine learning problems: it is a problem we will face in the future, not a problem that we face today. Nevertheless, it is crucial that we solve this problem \textit{before} it becomes serious, as even a single failure of superintelligence misalignment in practice could be catastrophic. %

This presents a major methodological challenge: how do we even approach studying a problem that is not yet a problem? How do we make progress on the core difficulties of superalignment? How do we make progress with today’s systems, knowing that our efforts will not be wasted by surprising new model capabilities that will inevitably arise in the future~\citep{wei2022emergent}?
We do not claim to have a complete answer to these questions, but we outline some best practices for maximizing our chances of making real progress on superalignment.

\textbf{Analogous setups.}\quad We should construct increasingly analogous empirical setups, and we should enumerate any remaining disanalogies. A setup is analogous if our results on that setup do not rely on assumptions that will break down in the future, making results today likely qualitatively similar to results in the future. Our main evaluation setup, introduced in \Cref{sec:methodology}, is intended to be more analogous to the superalignment problem. We enumerate some remaining disanalogies with our setup in \Cref{subsec:disanalogies}.

\textbf{Enumerating assumptions.}\quad We should enumerate the key assumptions that our results (either implicitly or explicitly) rely on. Clarifying what assumptions we are making makes it much easier to know when our results might break down. We enumerate our main disanalogies and assumptions in \Cref{subsec:disanalogies} and \Cref{sec:assumptions}. %

\textbf{Sensitivity analysis.}\quad We should evaluate the sensitivity of our results to changes in our assumptions and empirical setup. While we can make informed guesses about the future, we do not know exactly what future models will be like, so it is difficult to entirely trust any particular experimental setup. Validating that our results are robust to many different sets of assumptions can make us substantially more confident our results will transfer to the future superalignment problem. We do some initial sensitivity analysis in \Cref{sec:app-simulation}, and intend to do much more in future work.

\textbf{Scalable techniques.}\quad We should avoid techniques that rely on assumptions that will likely break down for future (superhuman) models. For example, when we do few-shot prompting we are intuitively incentivizing models to predict some useful distribution of human text, whereas when we do finetuning we are intuitively incentivizing a model to output what it knows regardless of how it knows it. This is one of the reasons we focus on finetuning methods in this paper: they are more likely to scale to superhuman models compared to prompting.

\textbf{Incidental usefulness today.}\quad 
One possible validation that progress on our setup is real would be to show that it is incidentally useful in practice today;
while we advocate focusing on the core challenges of superalignment, if our findings are never useful with today's models that would be evidence that we are not on the right track. One example of a near-term practical milestone would be to align GPT-4 on instruction-following tasks using only GPT-3-level supervision; if we could get strong alignment without any humans involved at all, that would make alignment much simpler and cheaper today. However, usefulness today is certainly not sufficient for aligning superintelligence, and in general a common failure mode of empirical alignment research is it prioritizes usefulness today at the expense of analogousness and scalability.

\textbf{Updating over time.}\quad We should update our evaluations and validate past findings as we learn more about what future models will look like. While we focus on the pretrained language model paradigm today, we plan on updating our setup if or when this stops being the dominant paradigm.

\section{How weak-to-strong generalization fits into alignment}
\label{sec:app-alignment}

Superintelligent AI systems will be extraordinarily powerful; humans could face catastrophic risks including even extinction~\citep{ai-risk-open-letter} if those systems are misaligned or misused.  
It is important for AI developers to have a plan for aligning superhuman models ahead of time---before they have the potential to cause irreparable harm.

Our plan for aligning superintelligence is a work in progress, but we believe that weak-to-strong techniques could serve as a key ingredient. 
In this section we sketch several illustrative possiblities for how we could use weak-to-strong generalization to help align superintelligent systems.%

\subsection{High-level plan}

\cite{superalignment} propose the following high level plan, which we adopt:
\begin{enumerate}
\item Once we have a model that is capable enough that it can automate machine learning---and in particular alignment---research, our goal will be to align that model well enough that it can safely and productively automate alignment research.
\item We will align this model using our most scalable techniques available, e.g.~RLHF~\citep{christiano2017deep,ouyang2022training}, constitutional AI~\citep{bai2022constitutional}, scalable oversight~\citep{saunders2022self,bowman2022measuring}, adversarial training, or---the focus of this paper----weak-to-strong generalization techniques.
\item We will validate that the resulting model is aligned using our best evaluation tools available, e.g.~red-teaming~\citep{perez2022red,perez2022discovering} and interpretability~\citep{ribeiro2016should,olah2018the,bills2023language,li2023inference}.
\item Using a large amount of compute, we will have the resulting model conduct research to align vastly smarter superhuman systems. We will bootstrap from here to align arbitrarily more capable systems.
\end{enumerate}

The goal of weak-to-strong generalization is to ensure step (2) is solved: align the first model capable of automating machine learning and alignment research. 
Importantly, this first model will likely be qualitatively superhuman along important dimensions, so RLHF is unlikely to be sufficient (\Cref{sec:results}). If we had a superhuman model, how would we apply weak-to-strong generalization to align it?

\subsection{Eliciting key alignment-relevant capabilities with weak-to-strong generalization}

There are many different alignment-relevant capabilities we could try to elicit from a superhuman model that could significantly help with alignment, including:\footnote{Ideally we elicit several related concepts and verify that we get consistent answers between them.}
\begin{itemize}
\item \textbf{Safety:} does a given behavior produced by an AI system risk the safety of human lives or well-being in important ways?
\item \textbf{Honesty:} is a given natural language statement true or false?
\item \textbf{Instruction following:} does a given behavior produced by an AI system follow a user's instruction faithfully?
\item \textbf{Code security:} does some given code have important security vulnerabilities or backdoors? Is it safe to execute it?
\end{itemize}

In the ideal case, the capability we elicit from the model would be robust enough that we can turn it into a reward model and safely optimize it; future work should assess the feasibility of this approach.
At the opposite extreme, we could potentially use the elicited capability as an ``oracle'' that we can manually query; 
intuitively, if we had a superhuman oracle model, we may be able to leverage it to help us bootstrap to a more robust alignment solution, even if that oracle is not itself entirely robust.

\subsection{Alignment plan assumptions}\label{sec:assumptions}

Many alignment plans which appear different on the surface actually depend on heavily correlated assumptions. For a given alignment plan, it is also often unclear which subproblems the plan attempts to solve, and which subproblems the plan assumes are unlikely to be an obstacle. As a result, we think enumerating assumptions is an important part of making progress on alignment.

In addition to the major disanalogies discussed in \Cref{subsec:disanalogies}, the assumptions we make for an alignment plan based on weak-to-strong generalization include:

\begin{itemize}
    \item \textbf{No deceptive alignment in base models.} 
    We assume that pretrained base models (or the equivalent in future paradigms) will be highly intelligent but not highly agentic (e.g.~will not have long-term goals)---and consequently will not be deceptively aligned~\citep{hubinger2019risks,ngo2022alignment,carlsmith2023scheming} out-of-the-box. Our goal is to elicit the superhuman capabilities of this capable but safe base model, and use those capabilities to create an aligned (possibly agentic) superhuman model.
    \item \textbf{Elicited concepts are sufficiently robust, or do not need to be.} We assume it is either possible to solve alignment using only a small amount of optimization applied to the capabilities we elicit, or that it is possible to make weak-to-strong elicited capabilities sufficiently robust against overoptimization.
    \item \textbf{The concepts we care about are natural to future AGI.} The superhuman base model we apply weak-to-strong generalization to has some ``alignment-complete'' concept, such as honesty, that is extrapolated in the way we would endorse if we could understand everything the superhuman model understands, and which is natural enough to the model that it is feasible to elicit. 
    \item \textbf{Sufficiently gradual takeoff.} Before we have superintelligence, we will have somewhat superhuman models long enough that we can use them to finish solving the full superintelligence alignment problem. 
    We can use it to solve superalignment before it causes recursive self improvement or catastrophic damage.
    \item \textbf{Moderately superhuman models are sufficient to solve alignment.} 
    We assume the first models capable of automating alignment research in practice are moderately superhuman, i.e. in a regime similar to what we study empirically in this work. For example, we may assume that we only need to bridge a weak-strong gap of at most (say) 4 OOMs of effective compute.
    \item \textbf{No need to solve human values.} 
    We assume we do not need to solve hard philosophical questions of human values and value aggregation before we can align a superhuman researcher model well enough that it avoids egregiously catastrophic outcomes.
\end{itemize}

This list represents a non-exhaustive set of notable assumptions we often operate under, and we will constantly reassess and update these assumptions over time as we learn more.
We \textit{do not} think these are necessarily valid assumptions by default, and believe it is important to validate them, work towards making them true, or mitigate failure modes from them being invalid.

Furthermore, there are a huge number of uncertainties about what future AI systems will look like and exactly how we should align them. 

\bibliographystyle{style/iclr2023_conference}

\end{document}